\documentclass[review,10pt]{JMtemplate}

\usepackage{bm}
\usepackage{graphicx} 
\usepackage{amsmath,mathrsfs,amssymb} 
\usepackage{amsfonts}  % blackboard math symbols, e.g., \mathbb
\usepackage{algorithm}
\usepackage{algorithmic}
\usepackage{siunitx}   % 用于书写 数字+单位 的格式
\usepackage{upgreek}   % 用于书写 数字+单位 的格式
\usepackage{xcolor}    % colors
\usepackage{subfigure}
\usepackage{booktabs}       % professional-quality tables
\usepackage{longtable}
\usepackage{threeparttable}
\usepackage{multirow}
\usepackage{multicol}
\usepackage{times}
\usepackage{helvet}
\usepackage{hyperref}
\usepackage{courier}
\usepackage{natbib}  % DO NOT CHANGE THIS AND DO NOT ADD ANY OPTIONS TO IT
\usepackage{caption} % DO NOT CHANGE THIS AND DO NOT ADD ANY OPTIONS TO IT

\newtheorem{theorem}{Theorem}
\newtheorem{definition}{Definition}

\newcommand*{\dif}{\mathop{}\!\mathrm{d}}

\begin{document}
\begin{frontmatter}
\title{Depth-induced NTK: Bridging Over-parameterized Neural Networks and Deep Neural Kernels}

\author{Yong-Ming Tian\textsuperscript{\rm 1,2}}
\author{Shuang Liang\textsuperscript{\rm 1,2}}
\author{Shao-Qun Zhang\textsuperscript{\rm 1,2,}\footnote{Shao-Qun Zhang is the corresponding author.}}
\author{Feng-Lei Fan\textsuperscript{\rm 3}}

\address{\textsuperscript{\rm 1}State Key Laboratory of Novel Software Technology, Nanjing University, China \\
	\textsuperscript{\rm 2}School of Intelligent Science and Technology, Nanjing University, China \\
	\textsuperscript{\rm 3}Department of Data Science, City University of Hong Kong, Hong Kong\\
    \{tianym, liangs, zhangsq\}@lamda.nju.edu.cn, hitfanfenglei@gmail.com}
\date{\today}

%\author{Zhi-Hua Zhou\corref{cor1}}
%\address{National Key Laboratory for Novel Software Technology\\
%Nanjing University, Nanjing 210093, China} \cortext[cor1]{\small Corresponding author.
%Email: zhouzh@nju.edu.cn}

\begin{abstract}
While deep learning has achieved remarkable success across a wide range of applications, its theoretical understanding of representation learning remains limited. Deep neural kernels provide a principled framework to interpret over-parameterized neural networks by mapping hierarchical feature transformations into kernel spaces, thereby combining the expressive power of deep architectures with the analytical tractability of kernel methods. Recent advances, particularly neural tangent kernels (NTKs) derived by gradient inner products, have established connections between infinitely wide neural networks and nonparametric Bayesian inference. However, the existing NTK paradigm has been predominantly confined to the infinite-width regime, while overlooking the representational role of network depth. To address this gap, we propose a depth-induced NTK kernel based on a shortcut-related architecture, which converges to a Gaussian process as the network depth approaches infinity. We theoretically analyze the training invariance and spectrum properties of the proposed kernel, which stabilizes the kernel dynamics and mitigates degeneration. Experimental results further underscore the effectiveness of our proposed method. Our findings significantly extend the existing landscape of the neural kernel theory and provide an in-depth understanding of deep learning and the scaling law.
\end{abstract}

\begin{keyword}
deep neural kernels \sep over-parameterized neural networks \sep weak dependence \sep training invariance \sep spectrum
\end{keyword}
\end{frontmatter}

\section{Introduction}\label{Introduction}
Deep learning has gained remarkable success in domains such as computer vision~\citep{krizhevsky2012imagenet}, natural language processing~\citep{devlin2019bert}, and scientific computation~\citep{raissi2019physics}. Despite these advances, the theoretical understanding of how deep neural networks learn and generalize remains incomplete~\citep{poggio2020theoretical}. In particular, the interplay between depth, over-parameterization, and representation learning poses fundamental challenges to classical statistical learning and existing deep learning theory~\citep{belkin2021fit,zhou2021over}. Recently, deep neural kernels provide a functional perspective that bridges deep neural networks and kernel methods by interpreting the hierarchical feature transformations in neural networks as implicit kernel mappings. Unlike conventional statistical kernels that rely on predefined similarity measures such as the Gaussian or polynomial kernels and the existing deep learning theory that focuses on the classification margin~\citep{bartlett2017spectrally, lyu2022improving} and function norm~\citep{neyshabur2015norm, zhang2017understanding}, deep neural kernels learn data-dependent similarity structures through multi-layer nonlinear transformations, especially by building the correspondence between over-parameterized neural networks and Gaussian processes~\citep{NIPS2016_abea47ba, Matthews2018GaussianPB, Neal1996}. This manner enables the combination of the expressive power of deep architectures with the theoretical interpretability of kernel methods. Representative neural kernels include the Neural Network Gaussian Process (NNGP)~\citep{lee2018:NNGP} and Neural Tangent Kernel (NTK)~\citep{WideNTK2021}, which are derived from the variable inner product in the forward-propagation process and the gradient inner product in the backward-propagation process of over-parameterized neural networks, respectively.

\begin{figure}[t]
	\centering
	\includegraphics[width=1\columnwidth]{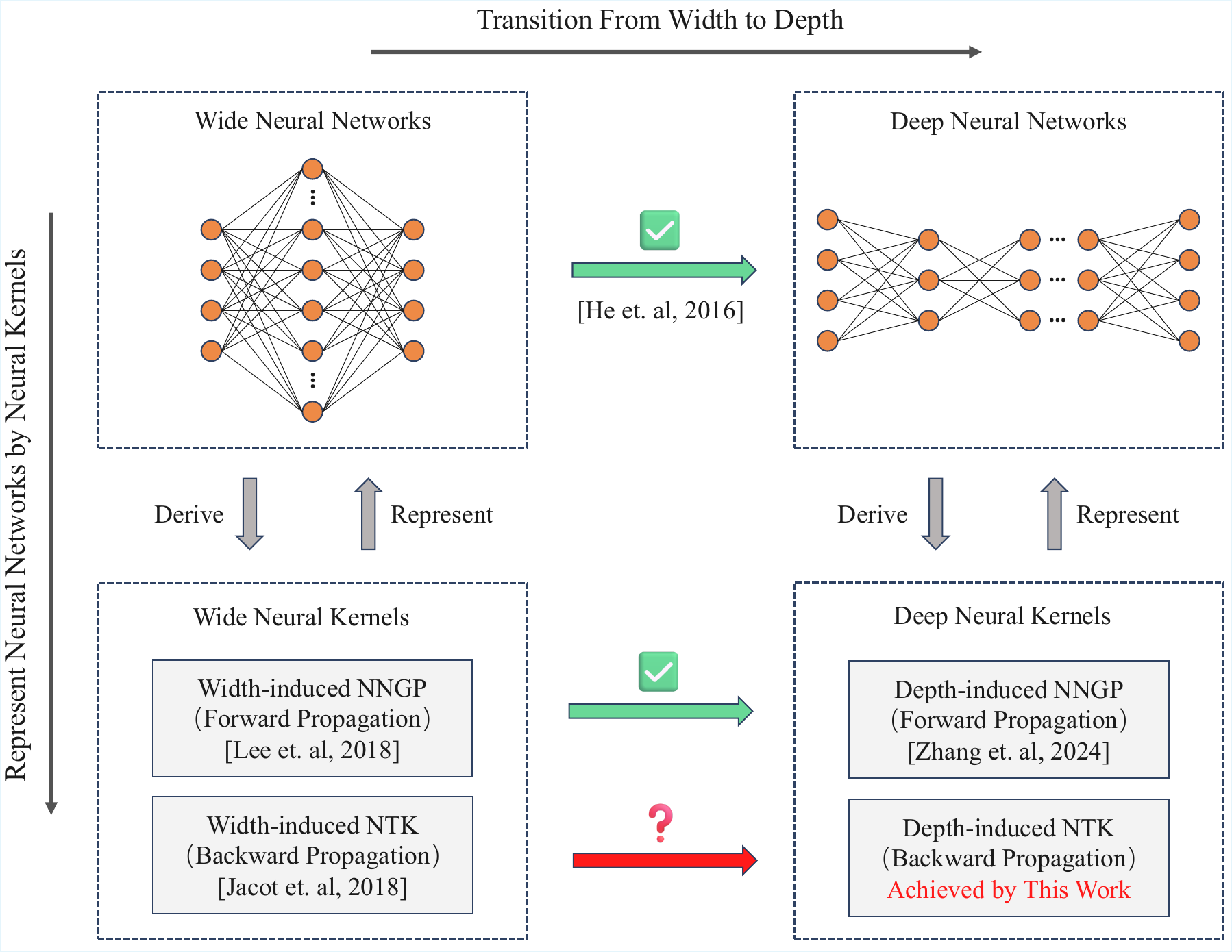}
	\caption{The transition of research focus from width to depth in both neural networks and neural kernels.}
	\label{fig:Transition_from_width_to_depth}
\end{figure}

Notably, most existing studies of NNGP and NTK kernels build upon the assumption that the network width of deep learning models approaches infinity, leaving the representational role of network depth insufficiently explored. As illustrated by Figure~\ref{fig:Transition_from_width_to_depth}, recent research focus has progressively shifted from width to depth, both in the study of neural networks and neural kernels. This transition is largely driven by the recognition that understanding the theory of over-parameterized deep learning representations fundamentally hinges on grasping the role of depth. Bengio et al.~\citep{bengio2013representation} have also pointed out that depth is the key to endowing over-parameterized neural networks with hierarchical and higher-order representation capabilities. Aligned with these recognitions, He et al.~\citep{he2016deep} have proposed residual neural networks, which leverage the key idea of shortcut connections to enable deeper layers. This marks the transition from increasing network width to increasing depth, significantly enhancing the representational power of neural networks. A similar transition has occurred in the study of neural kernels. More recently, researchers have developed deep neural kernels that are derived from infinitely deep neural networks, such as the depth-induced NNGP kernel, which successfully characterizes how depth influences the forward-propagation process of deep neural networks~\citep{Zhang2024:deep}.

Despite the encouraging advances in depth-induced NNGP, formulating a corresponding depth-induced NTK remains substantially more challenging. The difficulties can be attributed to two primary factors. (1) Existing width-induced NTK kernels are established under the assumptions of infinite width~\citep{kim2023infinite} or asymptotically increasing width~\citep{banerjee2023neural}. Under these conditions, the kernel is characterized by gradient inner products at an arbitrarily wide layer; thus the independence across neurons allows the Central Limit Theorem to apply directly, ensuring convergence. In contrast, the depth-induced NTK depends on gradients aggregated across a large number of layers to reveal how depth shapes representation. However, intrinsic inter-layer correlations break the independence structure, making the classical Central Limit Theorem inapplicable to the depth-induced NTK. (2) The layer-wise varying inner products that arise during forward propagation induce a weakly dependent sequence, for which a depth-induced NNGP can be derived via the Generalized Central Limit Theorem. However, the weak dependence structure identified in forward propagation cannot be transferred in a direct or equivalent manner, as the NTK kernel is defined through backward-propagation dynamics. These challenges collectively reveal a theoretical gap in understanding how network depth influences neural kernel representations beyond the infinite-width regime. Numerous researchers have argued that the infinite-width assumption fails to explain the success of deep learning~\citep{huang2020dynamics,seleznova2022neural}, resulting in a noticeable performance gap between kernel regression with NTK and deep neural networks. Consequently, extending the NTK framework toward a depth-induced neural kernel paradigm is of significant importance. Such a framework could reveal how depth fundamentally shapes representation learning during backward optimization of deep networks and potentially bridge the performance gap between NTK-based kernel regression and practical deep neural networks.

In this paper, we propose a depth-induced NTK kernel NTK$_{(d)}$, based on a shortcut-related network architecture that relates closely to residual neural networks~\citep{he2016deep}, which effectively enable deeper layers. In contrast to conventional neural kernels, the depth-induced NTK kernel, which is defined by an inner product of the combination of the parameter gradients of the shortcut-related layers and the parametric matrix, employs extra block diagonal matrices to scale the parameter gradients, leading to a weakly dependent sequence induced by network depth. We prove that the depth-induced NTK kernel converges to a Gaussian process as the network depth approaches infinity. Unlike the traditional width-induced NTK kernel NTK$_{(w)}$, which relies on infinitely wide neural networks, the depth-induced NTK kernel relies on infinitely deep neural networks. We theoretically characterize the training invariance of our proposed NTK$_{(d)}$ kernel and its spectrum by establishing considerably strict bounds for its smallest eigenvalue and largest singular value. Table~\ref{table: properties_summary_transposed} shows our progress compared to traditional neural kernels. 

\begin{table}[t]
	\centering
	\vspace{0.3em}
	\begin{tabular}{l|c|c|c}
		\toprule
		\textbf{Kernels} & \textbf{Existence} & \textbf{Spectrum} & \textbf{Training Invariance}   \\
		\midrule
		\multirow{2}{*}{\textbf{NNGP$_{(w)}$}} & Lee et al.~\citep{lee2018:NNGP} & \multirow{2}{*}{$\star\star$} & \multirow{2}{*}{--}   \\
		& Section 2 &  &    \\
		\midrule
		\multirow{2}{*}{\textbf{NNGP$_{(d)}$}} & Zhang et al.~\citep{Zhang2024:deep} & Zhang et al.~\citep{Zhang2024:deep} & \multirow{2}{*}{--}    \\
		& Theorem 1 & Theorem 3 &     \\
		\midrule
		\multirow{2}{*}{\textbf{NTK$_{(w)}$}} & Jacot et al.~\citep{WideNTK2021} & Nguyen et al.~\citep{pmlr-v139-nguyen21g} &  Jacot et al.~\citep{WideNTK2021}  \\
		& Theorem 1 & Theorem 3.2 &  Theorem 2   \\
		\midrule
		\multirow{2}{*}{\textbf{NTK$_{(d)}$}} & \multirow{2}{*}{Theorem~\ref{thm:NTK_depth}} & \multirow{2}{*}{Theorems~\ref{thm:smallest_eigenvalue} and~\ref{thm:largest_eigenvalue}} & \multirow{2}{*}{Theorem~\ref{thm:constant}}  \\
		&  &  &     \\
		\bottomrule
	\end{tabular}
	\caption{Key properties of our NTK$_{(d)}$ kernel compared to the NNGP$_{(w)}$, NNGP$_{(d)}$, and NTK$_{(w)}$ kernels, where $\star\star$ and $-$ denote no results so far and no need of the certain property, respectively.}
	\label{table: properties_summary_transposed}
\end{table}

Our main contributions can be summarized as follows:
\begin{itemize}
	\item {\bf Existence and Training Invariance of Depth-induced NTK.} We propose an NTK kernel derived by increasing depth based on a specific shortcut-related network architecture, which greatly broadens the scope of existing NTK theories to enable the analysis of deep neural networks. Its existence is guaranteed by Theorem~\ref{thm:NTK_depth} in Subsection~\ref{subsec: exist}. Based on its existence, we further prove that the depth-induced NTK kernel remains invariant during training under certain conditions, which allows the training dynamics of deep neural networks to be accurately characterized by kernel regression. We theoretically prove its training invariance by Theorem~\ref{thm:constant} in Subsection~\ref{subsec: theory_invariance}, and validate it by conducting experiments in Subsection~\ref{subsec: exp_invariance}.
	
	\item {\bf Spectrum of Depth-induced NTK.} We characterize the spectrum of the depth-induced NTK kernel. We have two significant findings: (1) The lower bound on the smallest eigenvalue of the depth-induced NTK kernel grows at least on the order of the dimension of the input data asymptotically, which guarantees that the depth-induced NTK kernel cannot degenerate to a zero kernel. Also, such a lower bound can be used to analyze the generalization property of deep neural networks. We theoretically prove the lower bound by Theorem~\ref{thm:smallest_eigenvalue} in Subsection~\ref{subsec: bound_it}, and validate it by conducting experiments in Subsection~\ref{subsec: lower_bound_exp}. (2) The upper bound on the largest singular value of the depth-induced NTK kernel grows at most quadratically with the number of shortcut connections of the shortcut-related network. The upper bound constrains the condition number of the depth-induced NTK kernel between any two inputs, which is associated with convergence behavior and training stability of neural networks. We theoretically prove the upper bound by Theorem~\ref{thm:largest_eigenvalue} in Subsection~\ref{subsec: bound_it}, and validate it by conducting experiments in Subsection~\ref{subsec: lower_bound_exp}.
	
	\item {\bf Experimental Validation of Depth-induced NTK.} We conduct experiments on benchmark data sets to verify the effectiveness of the proposed depth-induced NTK kernel, comparing it to the conventional width-induced NTK kernel. The experimental results are presented in Subsections~\ref{subsec:Kernel regression on sine function} and~\ref{subsec:Kernel Regression on Fashion-MNIST}. The experimental results suggest that the proposed depth-induced NTK kernel has comparable capability to the traditional width-induced NTK kernel. Besides, we investigate the influence of the separation constant $\hbar$ and the number of shortcut connections $K$ of the deep neural network on the characteristics of the depth-induced NTK kernel in Subsection~\ref{subsec:Effects of interval hbar and depth K on NTK(d)}. The experimental results suggest that the number of shortcut connections $K$ has a more significant impact on the characteristics of the depth-induced NTK kernel than the separation constant $\hbar$. This empirical finding is consistent with our theoretical results in Theorem~\ref{thm:NTK_depth}.
	
\end{itemize}

The rest of this paper is organized as follows. Section~\ref{Preliminary} introduces useful notations and the traditional width-induced NTK kernel, and the shortcut-related network architecture that is used to induce an NTK kernel by increasing depth. Section~\ref{Main Results} formally presents the depth-induced NTK kernel, giving the lower bound on its smallest eigenvalue and the upper bound on its largest singular value, and proving that it remains invariant during training. Section~\ref{Experiments} conducts numerical experiments. Section~\ref{Conclusions} concludes our work with prospects.

\section{Preliminaries}\label{Preliminary}
This section introduces useful notations, reviews the conventional width-induced NTK kernel, and provides a shortcut-related network architecture that is employed to derive an NTK kernel by increasing depth.
\subsection{Notations}\label{Notations}
Let $[N] = \{1,2,\dots,N\}$ be the set for an integer $N > 0$. Given two functions $g(n)$ and $h(n)$, we denote by $h(n)=\mathcal{O}(g(n))$ if there exist positive constants $c$ and $n_0$ such that $h(n) \leq cg(n)$ for every $n \geq n_0$; $h(n) = \Omega(g(n))$ if there exist positive constants $c$ and $n_0$ such that $h(n) \geq cg(n)$ for every $n \geq n_0$; $h(n) = \Theta(g(n))$ if there exist positive constants $c_1, c_2$ and $n_0$ such that $c_1g(n) \leq h(n) \leq c_2g(n)$ for every $n \geq n_0$. For a neural network, we employ $\phi$ to denote the element-wise activation and use $\dot{\phi}$ to denote its derivative. We employ $n_{l}$ to denote the number of neurons in the $l$-th layer. Provided a vector $\boldsymbol{x}\in\mathbb{R}^{n\times 1}$, we use $\text{vec}(\boldsymbol{x}_i)_{i=1}^n$ to denote the component form of the column vector $\boldsymbol{x}$ and use $\text{diag}\{\dot{\phi}(\boldsymbol{x})\}$ to denote the diagonal matrix where the diagonal entries are $\dot{\phi}(x_1), \dot{\phi}(x_2)\cdots \dot{\phi}(x_n)$. For a matrix $\mathbf{W}\in\mathbb{R}^{n\times m}$, we denote its eigenvalues by $\lambda$ and its singular values by $\sigma$. Let $\| \mathbf{W} \|_s$ denote the spectral norm for $\mathbf{W}$, that is, $\|\mathbf{W}\|_s \triangleq \sigma_{\max}(\mathbf{W})$, where $\sigma_{\max}(\mathbf{W})$ denotes the largest singular value of $\mathbf{W}$. Correspondingly, $\lambda_{\min}(\mathbf{W})$ denotes the smallest eigenvalue of $\mathbf{W}$. Let $\text{D}_b^{n}{\{{\mathbf{W}}\}}$ denote the $n\times n$ block diagonal matrix whose diagonal blocks are all $\mathbf{W}$. Let $\mathbf{W} \sim \mathcal{N}(\mu, \sigma^2)$ denote that each element of $\mathbf{W}$ is independently drawn from a Gaussian distribution with mean $\mu$ and variance $\sigma^2$. Given two matrices $\mathbf{A},\mathbf{B}\in\mathbb{R}^{n\times m}$, we denote by $\mathbf{A}\otimes\mathbf{B}$ their Kronecker product, and by $\mathbf{A}\circ\mathbf{B}$ their Hadamard product. For a sequence $\{A^0,A^1,A^2,\cdots,A^n\}$, we denote it as $\{A^i\}_{i = 0}^n$.

\begin{definition} \label{def:well_posed}
	An element-wise activation $\phi:\mathbb{R}\to\mathbb{R}$ is said to be \textbf{well-posed}, if $\phi$ is first-order differentiable, with its derivative bounded by a constant $C_{\phi}$. For a well-posed activation function $\phi$, a weight matrix $\mathbf{W}$ is said to be \textbf{stable-pertinent}, if the inequality $ C_{\phi} \|\mathbf{W}\|_s \leq \epsilon$ holds, where $\epsilon$ is a constant that satisfies $0 < \epsilon < 1$.
\end{definition}
Definition~\ref{def:well_posed} restricts the activation and weights concerning a neural network. Many commonly used activation functions satisfy the well-posed property, such as ReLU, Leaky ReLU and Tanh, whose derivatives are bounded by $1$, and Sigmoid, bounded by $1/4$. As for the stable-pertinent property, the constant $\epsilon$ can be any small number, and is not necessarily required to be less than 1; the restriction $\epsilon<1$ is imposed merely to simplify some of the proofs of the main theorems in this paper. We claim that the stable-pertinent property is commonly used, which assumes that the norm of weights is upper-bounded. For example, Jiao et al.~\citep{jiao2023approximation} studied ReLU neural networks with upper-bounded norm constraints on weights. Previous works have claimed that this property is beneficial for controlling the generalization ability of neural networks~\citep{bartlett2017spectrally, golowich2018size}.

\begin{definition} \label{def: well_scaled} A data distribution $P_X$ is said to be \textbf{well-scaled}, if the followings hold
	\begin{itemize}
		\item[(1)] $\int \boldsymbol{x}\,\dif P_X(\boldsymbol{x}) = 0,$
		\item[(2)] $\int \left\|\boldsymbol{x}\right\|_2\dif P_X(\boldsymbol{x}) = \Theta(\sqrt{d})$,
		\item[(3)] $\int \left\|\boldsymbol{x}\right\|_2^2 \dif P_X(\boldsymbol{x}) = \Theta(d)$,
		\item[(4)] the data distribution $P_X$ satisfies the Lipschitz concentration property, that is, given any Lipschitz continuous function $f: \mathbb{R}^d \rightarrow \mathbb{R}$ with Lipschitz constant $L$, for $\forall~\delta>0$, there exists a positive constant $c > 0$, such that
		\[\mathbb{P}\left(\left|f(\boldsymbol{x}) - \int f(\boldsymbol{x}^\prime) \dif P_X (\boldsymbol{x}^\prime)\right| > \delta \right) \leq 2\exp({{-c\delta^2}/{L^2}})\ .\]
	\end{itemize}
\end{definition}
Definition~\ref{def: well_scaled} imposes restrictions on the data distribution, which have served as conditions to characterize the spectrum of the width-induced NTK~\citep{simon2019gradient, pmlr-v139-nguyen21g}. The conditions on the data distribution are mild. The first three conditions in Definition~\ref{def: well_scaled} simply concern the scaling of the data vector $\boldsymbol{x}$. The fourth condition accommodates a broad class of distributions that satisfy the log-Sobolev inequality, such as the standard Gaussian distribution, the uniform distribution on the sphere or on the unit hypercube~\citep{pmlr-v139-nguyen21g}. Throughout the paper, we consider data points $\left\{\boldsymbol{x}_i\right\}_{i=1}^{N}$ i.i.d. drawn from a centered Gaussian distribution $P_X$, which is well-scaled.

\subsection{Width-induced Neural Tangent Kernel}\label{Neural Tangent Kernel}
Considering an $L$-layer neural network $f:\mathbb{R}^d \to \mathbb{R}^{n_o}$, we use $\mathbf{W}$ to denote the variable of weight parameters where $\mathbf{W}^l$ denotes that of the $l$-th layer, and neglect the biases $\boldsymbol{b}$ for simplicity. The feed-forward propagation of the concerned neural network follows
\begin{equation}\label{eq:forward}
	\boldsymbol{z}^{0} = \boldsymbol{x} \ , \quad
	\boldsymbol{z}^{l} = \phi\left(\frac{1}{\sqrt{n_l}}\mathbf{W}^{l} \boldsymbol{z}^{l-1}\right)\quad \text{for} \quad l \in [L] \ .
\end{equation}
Notice that we normalize the weight matrices $\mathbf{W}^l$ by a scaling factor $1/\sqrt{n_l}$ for $l \in [L]$, a technique that facilitates the proof of the invariance of NTK during training in Theorem~\ref{thm:constant}, which is consistent with the conventional study of Jacot et al.~\citep{WideNTK2021}. The operations of feed-forward and back-ward propagation using the initialization of \( \mathbf{W}^l \sim \mathcal{N}(0, 1) \) and the additional scaling factor $1/\sqrt{n_l}$ are consistent with those using the initialization of \( \mathbf{W}^l \sim \mathcal{N}(0, 1/\sqrt{n_l}) \) without the explicit scaling. For simplicity, we use the symbol $\hat{\mathbf{W}}^l$ to denote $\mathbf{W}^l/\sqrt{n_l}$ for $l \in [L]$ throughout the paper. 

The output of the network is computed by $f(\boldsymbol{x}) = \boldsymbol{z}^{L}$. The objective is to learn $\boldsymbol{y} = f(\boldsymbol{x})$, supervised by the loss function $\mathcal{L}(f(\boldsymbol{x}),\boldsymbol{\hat{y}}) = \|f(\boldsymbol{x})-\boldsymbol{\hat{y}}\|_2^2$, where $\boldsymbol{\hat{y}}$ denotes the true label. Based on gradient descent, the dynamics of parameter updating can be described as follows
\[
\frac{\partial \mathbf{W}}{\partial t} = - \frac{\partial \mathcal{L}(f(\boldsymbol{x}),\boldsymbol{\hat{y}})}{\partial \mathbf{W}} = - \frac{\partial \mathcal{L}(f(\boldsymbol{x}),\boldsymbol{\hat{y}})}{\partial f(\boldsymbol{x})} \frac{\partial f(\boldsymbol{x})}{\partial \mathbf{W}} \ .
\]
Given two inputs $\boldsymbol{x}$ and $\boldsymbol{x}^\prime$, the width-induced NTK kernel NTK$_{(w)}$ is defined as the inner product of the gradients of parameters
\begin{equation}\label{eq:NTK_width_definition}
	\text{NTK}_{(w)}(\boldsymbol{x},\boldsymbol{x}^\prime) \triangleq \sum_{l=1}^{L} \left\langle\frac{\partial f(\boldsymbol{x})}{\partial \mathbf{W}^{l}}, \frac{\partial f(\boldsymbol{x}^\prime)}{\partial \mathbf{W}^{l}}\right\rangle \ .
\end{equation}
There is a claim that as the network width goes to infinity, that is, $n_l\rightarrow+\infty$ for $l\in[L]$, the dynamics of randomly initialized neural networks led by gradient descent can be equivalently described by kernel methods~\citep{WideNTK2021}, that is,
\[
\frac{\partial f(\boldsymbol{x})}{\partial t} = - \frac{1}{n}\sum_{i=1}^n \text{NTK}_{(w)}(\boldsymbol{x},\boldsymbol{x}_i) \frac{\partial \mathcal{L}(f(\boldsymbol{x}),\boldsymbol{\hat{y}})}{\partial f(\boldsymbol{x}_i)} \ ,
\]
where $n$ is the number of samples and $\boldsymbol{x}_i$ denotes the $i$-th training sample for $i \in [n]$.

Based on this recognition, the convergence and generalization properties of fully connected neural networks were studied~\citep{WideNTK2021}, and subsequent works extended the NTK results to other architectures, such as graph neural networks~\citep{du2019graph}, residual networks~\citep{huang2020deep}, convolutional networks~\citep{li2019enhanced}, and even attention-based models~\citep{Wu2023:encoder, Zhou2024:transformer}.  Later works further attempt to apply this width-induced NTK to the analysis of deep neural networks; however, they either observe a notable degradation in performance~\citep{huang2020dynamics,seleznova2022neural} or require impractical and unnatural initialization schemes to achieve standard performance~\citep{lee2022neural}.

\subsection{Shortcut-related Network Architecture}\label{subsec: Topology of Neural Network}

\begin{figure}[!htb]
	\centering
	\includegraphics[width=\columnwidth]{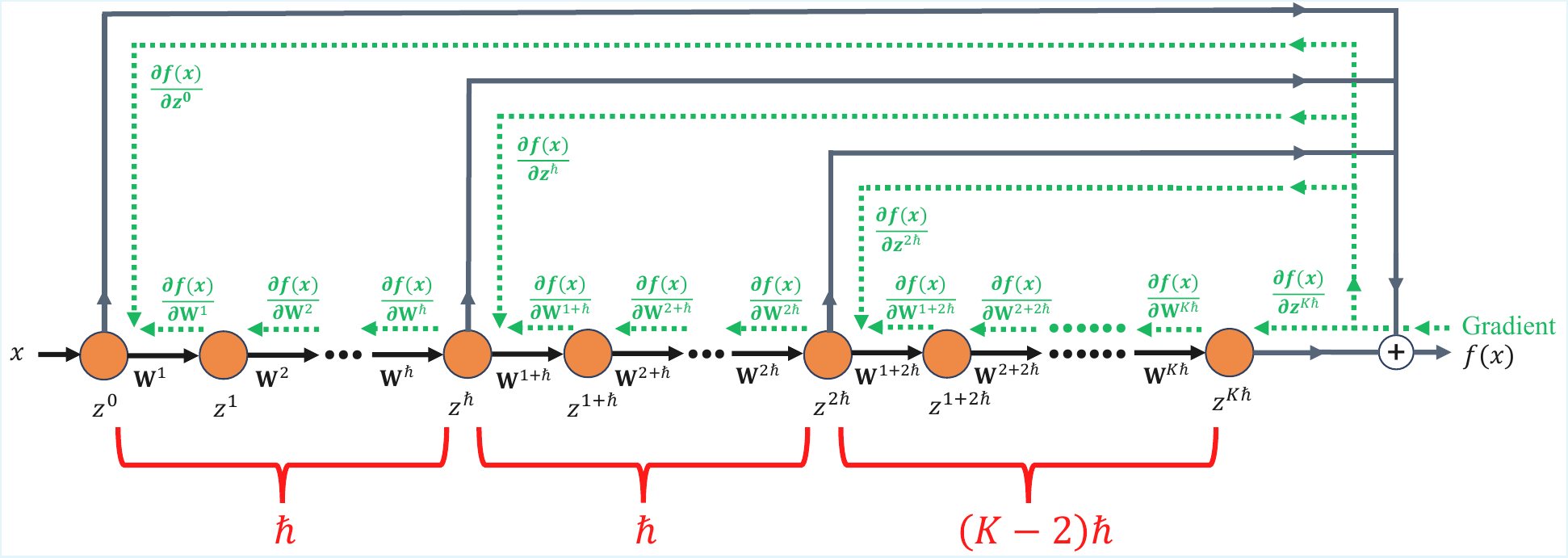}
	\caption{Illustration of the shortcut-related network architecture that derives an NTK by increasing depth.}
	\label{Figure_topology}
\end{figure}

This work considers a shortcut-related network with $L$ layers, whose architecture is illustrated in Figure~\ref{Figure_topology}. Formally, the feed-forward propagation of this network also follows Eq.~\eqref{eq:forward}, while the final output of this network $f:\mathbb{R}^d \to \mathbb{R}^{n_o}$ is computed by a mean of the outputs of preceding $K+1$ ($K\in\mathbb{N}^+$) separated layers
\begin{equation} \label{eq:shortcut}
	f(\boldsymbol{x}) = \frac{1}{\sqrt{M_{\boldsymbol{z}}}} \sum_{\kappa=0}^{K} \mathbf{J}^{\kappa\hbar} \boldsymbol{z}^{\kappa\hbar}(\boldsymbol{x}) \ ,
\end{equation}
where $\hbar\in\mathbb{N}^+$ is the separation constant, the all-ones matrix $\mathbf{J}^{\kappa\hbar} = [1]_{n_o \times n_{\kappa\hbar}}$ employed for dimensional normalization indicates the unit shortcut connection weights between $\boldsymbol{z}^{\kappa\hbar}$ and the output, and $M_{\boldsymbol{z}} = \sum\nolimits_{\kappa=0}^{K} n_{\kappa\hbar}$ is a scaling constant. In contrast to the fully connected architecture, where connection weights only bridge adjacent layers and the residual architecture that incorporates shortcut connections between hidden layers, the proposed architecture marshals the separated hidden layers to the final layer via shortcut connections.

The architecture employed in this paper has been effectively applied in multiple domains, including computer vision~\citep{fan2018sparse} and medical imaging~\citep{kang2019cycle, you2019ct}. A salient characteristic of this architecture is the incorporation of shortcut connections, inspired by residual networks~\citep{he2016deep}. Only the layers featuring such shortcut connections contribute directly to the final output, allowing information to propagate across multiple layers with minimal attenuation, whereas each of them is separated by $\hbar - 1$ intermediate layers without shortcut connections. While less mainstream, this architecture offers distinct advantages: as shown in Theorem~\ref{thm:NTK_depth} in Subsection~\ref{subsec: exist}, the shortcut architecture ensures the weak dependence among layer variables, which is crucial for applying the Generalized Central Limit Theorem and deriving the associated neural kernel. Looking ahead, the proposed methodology is expected to generalize to a variety of architectures, with its feasibility and generality supported by both theoretical analysis and empirical observations presented in this work.

\section{Main Results}\label{Main Results}
\begin{figure}[!htb]
	\centering
	\includegraphics[width=0.95\columnwidth]{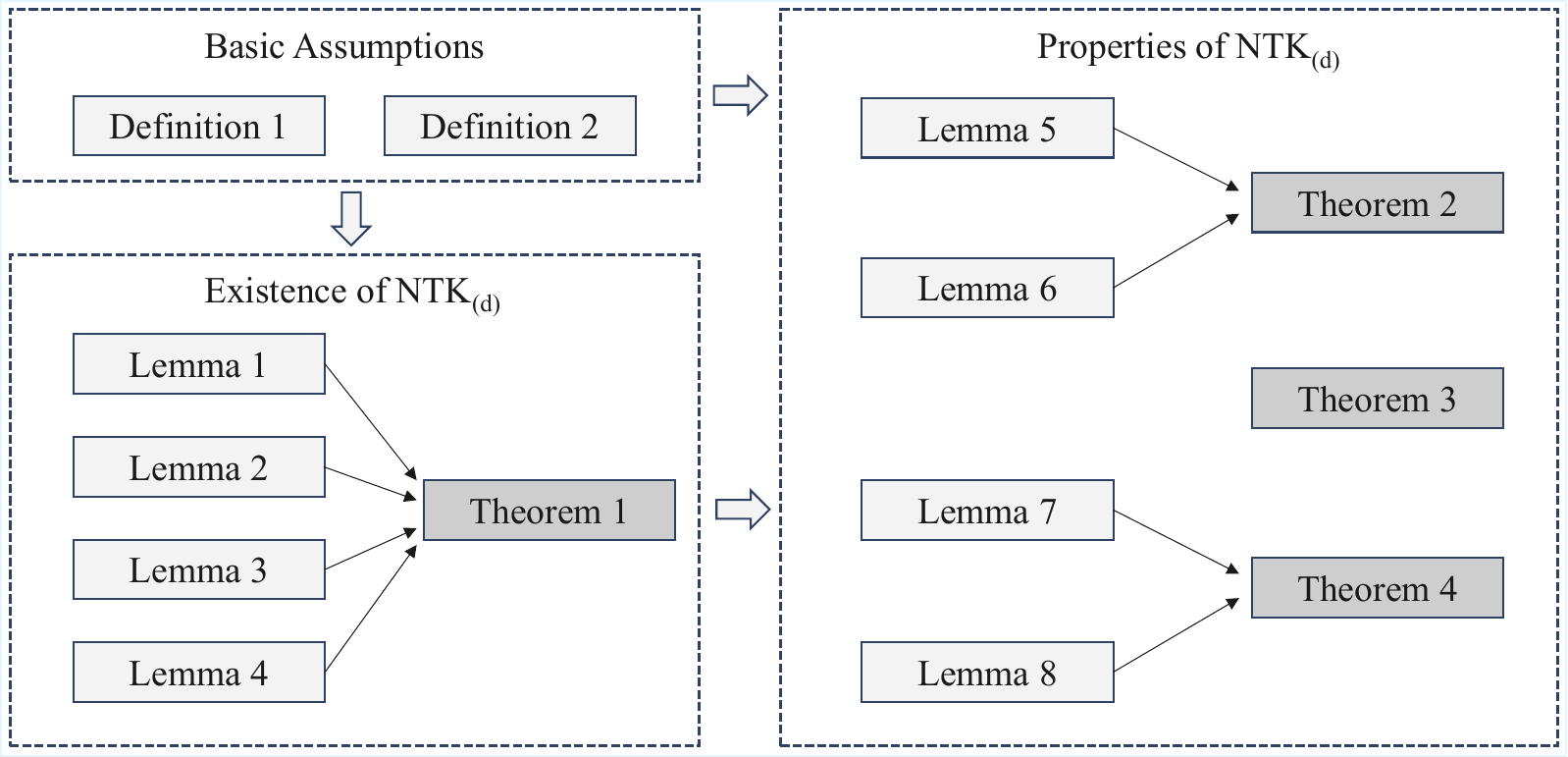}
	\vspace{-10pt}
	\caption{The schematic diagram of all the theoretical results in this paper.}
	\vspace{-12pt}
	\label{fig:schematic_diagram}
\end{figure}

\vspace{-10pt}
This section formally presents the depth-induced NTK kernel NTK$_{(d)}$ and its properties. Figure~\ref{fig:schematic_diagram} shows the schematic diagram of all the theoretical results in this paper.

\subsection{Existence of the Depth-induced NTK Kernel}\label{subsec: exist}
Now, we present our first theorem as follows.
\begin{theorem}\label{thm:NTK_depth}(Existence of Depth-induced NTK) Provided the shortcut-related network defined by Eqs.~\eqref{eq:forward} and~\eqref{eq:shortcut}, if the following conditions hold
	\begin{itemize}
		\vspace{-8pt}
		\item[(1)] the activation $\phi$ is well-posed,
		\vspace{-5pt}
		\item[(2)] the weight matrices are stable-pertinent for $\phi$, that is, $\mathbf{W}^l \in SP(\phi)$ for $\forall~ l\in [L]$,
	\end{itemize}
	\vspace{-10pt}
	there derives an NTK kernel that converges to a Gaussian distribution in the limit of the number of shortcut connections as well as the network depth going to infinity, that is, $K\rightarrow +\infty$ as well as $L\rightarrow +\infty$. Formally, for any inputs $\boldsymbol{x}$ and $\boldsymbol{x}^\prime$, we have
	\begin{equation}\label{main_NTK_sum}
		\text{NTK}_{(d)} (\boldsymbol{x}, \boldsymbol{x}') = \sum_{\kappa^\prime = 1}^{K}\text{NTK}_{(d)}^{\kappa^\prime\hbar} (\boldsymbol{x}, \boldsymbol{x}') \ ,
	\end{equation}
	with
	\begin{equation}\label{main_NTK_def}
		\text{NTK}_{(d)}^{\kappa^\prime\hbar} (\boldsymbol{x}, \boldsymbol{x}') 
		\triangleq \left\langle\text{D}_b^{n_{\kappa^\prime\hbar-1}} \{{\mathbf{W}^{\kappa^\prime\hbar}}^\top \}\frac{\partial f(\boldsymbol{x})}{\partial \mathbf{W}^{\kappa^\prime\hbar}}, \text{D}_b^{n_{\kappa^\prime\hbar-1}}\{{\mathbf{W}^{\kappa^\prime\hbar}}^\top\}\frac{\partial f(\boldsymbol{x}^\prime)}{\partial \mathbf{W}^{\kappa^\prime\hbar}}\right\rangle
	\end{equation}
	and
	\begin{equation}\label{main_NTK_exp}
		\text{NTK}_{(d)}^{\kappa^\prime\hbar} (\boldsymbol{x}, \boldsymbol{x}')= \frac{1}{{M_{\boldsymbol{z}}}}\left\langle\boldsymbol{z}^{\kappa^\prime\hbar-1}(\boldsymbol{x}),\boldsymbol{z}^{\kappa^\prime\hbar-1}(\boldsymbol{x}^\prime)\right\rangle \left\langle\Delta^{\kappa^\prime\hbar}(\boldsymbol{x}),\Delta^{\kappa^\prime\hbar}(\boldsymbol{x}^\prime)\right\rangle\ ,
	\end{equation}
	where 
	\[
	\Delta^{\kappa^\prime\hbar}(\boldsymbol{x}) = \sum_{\kappa = \kappa^\prime}^{K}\left[\prod_{i=\kappa^\prime\hbar}^{\kappa\hbar}\left(\hat{\mathbf{W}}^{i}\right)^\top\mathbf{D}^{i}(\boldsymbol{x})\right]~{\mathbf{J}^{\kappa\hbar}}^\top
	\quad\text{and}\quad
	\mathbf{D}^i(\boldsymbol{x}) = \text{diag}\{\dot{\phi}(\hat{\mathbf{W}}^i\boldsymbol{z}^{i-1}(\boldsymbol{x}))\} \ .
	\]
\end{theorem}
Theorem~\ref{thm:NTK_depth} states the existence of the NTK kernel NTK$_{(d)}$ induced by the network depth, which converges to a Gaussian distribution in the limit of $K\rightarrow +\infty$ as well as $L\rightarrow +\infty$, under mild constraints on activation and weight matrices.  In contrast to the traditional width-induced kernel NTK$_{(w)}$, which is based on the inner products of parameter gradients across all layers, our proposed depth-induced kernel NTK$_{(d)}$ employs extra block diagonal matrices to scale the parameter gradients. Also, it is only related to the layers with shortcut connections, which contribute to the output more directly. Moreover, we show that the NTK$_{(d)}^{\kappa^\prime\hbar}$ kernel can be expanded as the product of the inner product of $\boldsymbol{z}$ and that of $\Delta$, which enables a more straightforward analysis and contributes to the proofs of Theorems~\ref{thm:largest_eigenvalue} and~\ref{thm:constant}. The full proof of Theorem~\ref{thm:NTK_depth} can be accessed in Appendix~\ref{fullproof:3.1}.

The key idea of proving Theorem~\ref{thm:NTK_depth} is to exploit the Generalized Central Limit Theorem~\citep{b1995:clt}, where the sum of variables in a weakly dependent sequence converges to a Gaussian distribution. Thus, it suffices to prove that the sequence $\{\text{NTK}_{(d)}^{\kappa^\prime\hbar}\}_{\kappa^\prime=1}^{K}$ is weakly dependent. Zhang et al.~\citep{Zhang2024:deep} have proved that the feed-forward procedure of deep neural networks with apposite initialization derives a weakly dependent sequence. Our work extends this idea to $\{\text{NTK}_{(d)}^{\kappa^\prime\hbar}\}_{\kappa^\prime=1}^{K}$ by imposing some mild constraints on the activation and connection weights. Here, we start the proof sketch with some useful lemmas.
\begin{lemma} \label{lemma:main_weak}
	~\citep[Lemma 1]{Zhang2024:deep}  A random variable sequence $\{A^i\}_{i=1}^t$ can derive a $\beta$-mixing subsequence $\{A^{i\hbar}\}_{i=1}^{K}$ as $K\rightarrow +\infty$, where $\hbar\in\mathbb{N}^+$ is a sufficiently large number, if there exists a pair of positive constants $C_1,C_2$, such that for any $l\in\mathbb{N}$ satisfying $1 \leq l+\hbar \leq t$, one of the following conditions holds
	\[
	\left.
	\begin{aligned}
		(1)\quad&\left\|\frac{\partial A^{l+\hbar}}{\partial A^l}\right\|_s \leq C_1^\hbar\quad \text{and}\quad \frac{\left|A_p^{l+\hbar}A_q^{l}\right|}{\left|A_q^{l}\right|} \leq R~ C_1^\hbar\ ,  \\
		(2)\quad&\left\|\frac{\partial A^l}{\partial A^{l+\hbar}}\right\|_s \leq C_2^\hbar\quad \text{and}\quad \frac{\left|A_q^{l}A_p^{l+\hbar}\right|}{\left|A_p^{l+\hbar}\right|} \leq R~ C_2^\hbar\ ,
	\end{aligned}
	\right.
	\]
	where $A_p^{l+\hbar}$ and $A_q^{l}$ separately denote any element in $A^{l+\hbar}$ and $A^{l}$, and $R > 0$ is a statistical variable related to the $\beta$-mixing subsequence.
\end{lemma}
\vspace{-8pt}
Lemma~\ref{lemma:main_weak} gives a way to construct a $\beta$-mixing sequence, which is weakly dependent.

\begin{lemma}\label{lemma:main_prod_sum_weak}
	Let $\{X^i\}_{i=1}^{t}$ and $\{Y^i\}_{i=1}^{t}$ be two weakly dependent sequences constructed from $\{\tilde{X}^i\}_{i=1}^{t^\prime}$ and $\{\tilde{Y}^i\}_{i=1}^{t^\prime}$. We define $\tilde{A}^i = {\tilde{X}^i}\cdot{\tilde{Y}^i}$ and $\tilde{B}^i = {\tilde{X}^i}+{\tilde{Y}^i}$ for $i\in[t^\prime]$. Then, given the independence between $\tilde{X}^i$ and $\tilde{Y}^i$ for any $i\in[t^\prime]$, one has that the sequences $\{A^i\}_{i=1}^{t}$ and $\{B^i\}_{i=1}^{t}$ are also weakly dependent sequences constructed from $\{\tilde{A}^i\}_{i=1}^{t^\prime}$ and $\{\tilde{B}^i\}_{i=1}^{t^\prime}$, where $A^i = X^i\cdot Y^i$ and $B^i = X^i + Y^i$.
\end{lemma}
\vspace{-8pt}
Lemma~\ref{lemma:main_prod_sum_weak} states that the sum or product of two weakly dependent sequences remains weakly dependent, provided that the two sequences are independent of each other. The proof of Lemma~\ref{lemma:main_prod_sum_weak} can be found in Appendix~\ref{appendix:proof_of_prod_sum_weak}.

\begin{lemma}\label{weakCLT}
	~\citep[Theorem 27.4]{b1995:clt} Suppose that 
	\begin{itemize}
		\vspace{-8pt}
		\item[(1)] $\{A^i\}_{i=1}^n$ is stationary and $\alpha$-mixing with $\alpha_n = \mathcal{O}(n^{-5})$,
		\item[(2)] for $ \forall i \in [n]$, we have $\mathbb{E}[A^i] = 0$ and $\mathbb{E}[(A^i)^2] < \infty$.
		\vspace{-8pt}
	\end{itemize}
	Let $S_{n} = A^1 + A^2 + \dots + A^n$, then the following limit exists
	\[
	\rho^2 = \lim\limits_{n\to\infty} \frac{\mathbb{E}(S_n^2)}{n} \ .
	\]
	Further, given $\rho \neq 0$, the limit variable ${S_n}/({\rho\sqrt{n}})$ converges in distribution to the standard normal distribution, that is, ${S_n}/({\rho\sqrt{n}}) \overset{\mathrm{d}{}}{\to} \mathcal{N}(0,1)$ as $n\to\infty$.
\end{lemma}
\vspace{-8pt}
Lemma~\ref{weakCLT} shows the Generalized Central Limit Theorem for weakly dependent variables.

{\bf Finishing the Proof of Theorem~\ref{thm:NTK_depth}.} Our goal is to prove the weak dependence of the sequence $\{\text{NTK}_{(d)}^{\kappa^\prime\hbar}\}_{\kappa^\prime=1}^{K}$. By substituting Eq.~\eqref{eq:shortcut} into Eq.~\eqref{main_NTK_def}, we can derive the expanded form of NTK$_{(d)}$, which is composed by inner product of $\boldsymbol{z}$ and that of $\Delta$. The detailed proof of this step can be found in Appendix~\ref{subsec:expanded_form_of_delta}.

By the expanded form of NTK$_{(d)}^{\kappa^\prime\hbar}$ and Lemma~\ref{lemma:main_prod_sum_weak}, it suffices to prove that the sequences 
\[\{\boldsymbol{z}^{\kappa^\prime\hbar-1}(\boldsymbol{x})\}_{\kappa^\prime=1}^{K}\quad \text{and}\quad \{\Delta^{\kappa^\prime\hbar}(\boldsymbol{x})\}_{\kappa^\prime=1}^{K}\] are both weakly dependent. For the former, we prove it is a weakly dependent subsequence constructed from $\{\boldsymbol{z}^l\}_{l=0}^{L}$ by Lemma~\ref{lemma:main_weak}. From Eq.~\eqref{eq:forward}, we have $\boldsymbol{z}^l = \phi({\hat{\mathbf{W}}}^l {\boldsymbol{z}}^{l-1})$ for all $l \in [L]$. Given the well-posed $\phi$ and stable-pertinent parameter matrices, we have $\phi$ is first-order differentiable, with its derivative bounded by a certain constant $C_{\phi}$, and the inequality $ C_{\phi} \|\mathbf{W}\|_s \leq \epsilon$ holds, where $\epsilon$ is a constant that satisfies $0 < \epsilon < 1$. Thus, the conditions of Lemma~\ref{lemma:main_weak} are obviously satisfied well as
\[
\left\|\frac{\partial \boldsymbol{z}^{l+\hbar}}{\partial \boldsymbol{z}^l}\right\|_s \leq \epsilon^\hbar \quad\text{and}\quad \frac{| \boldsymbol{z}_p^{l+\hbar}\boldsymbol{z}_q^l|}{|\boldsymbol{z}_q^l|} \leq |\boldsymbol{z}_q^l|~ \epsilon^\hbar\ , 
\]
where $\boldsymbol{z}_p^{l+\hbar}$ and $\boldsymbol{z}_q^l$ separately denote the $p$-th element of $\boldsymbol{z}^{l+\hbar}$ and the $q$-th element of $\boldsymbol{z}^l$. Thus, the sequence $\{\boldsymbol{z}^{\kappa^\prime\hbar-1}(\boldsymbol{x})\}_{\kappa^\prime=1}^{K}$ is weakly dependent.

Secondly, for the sequence $\{\Delta^{\kappa^\prime\hbar}(\boldsymbol{x})\}_{\kappa^\prime=1}^{K}$ , by repeatedly employing Lemma~\ref{lemma:main_prod_sum_weak}, it suffices to prove that $\{A^{\kappa^\prime\hbar}\}_{\kappa^\prime=1}^K$ is weakly dependent, where
\[A^{\kappa^\prime\hbar} = \prod_{i=\kappa^\prime\hbar}^{K\hbar}\left(\hat{\mathbf{W}}^{i}\right)^\top\mathbf{D}^{i}(\boldsymbol{x})\ .\] 
Similarly, we verify the two conditions in Lemma~\ref{lemma:main_weak}. By the conditions in Theorem~\ref{thm:NTK_depth}, we can ensure that the following inequalities hold
\[
\left\|\frac{\partial A^{\kappa^\prime\hbar}}{\partial A^{\kappa^\prime\hbar+\hbar}}\right\|_s \leq \epsilon^{\hbar}
\quad\text{and}\quad
\frac{|A_{pq}^{\kappa^\prime\hbar}|_{\max} ~|A_{ij}^{\kappa^\prime\hbar+\hbar}|_{\max}}{|A_{ij}^{\kappa^\prime\hbar+\hbar}|_{\max}} \leq \left({\left(R_d\right)}^{3/2} |A_{ij}^{\kappa^\prime\hbar+\hbar}|_{\max}\right) ~\epsilon^{\hbar} \ , 
\] 
where $A_{pq}^{\kappa^\prime\hbar}$ and 
$A_{ij}^{\kappa^\prime\hbar+\hbar}$ separately denote the $(p,q)$-th element of $A^{\kappa^\prime\hbar}$ and the $(i,j)$-th element of $A^{\kappa^\prime\hbar+\hbar}$, and $R_d = \max\{n_{\kappa^\prime\hbar-1},n_{\kappa^\prime\hbar+\hbar-1},n_{K\hbar}\}$ is a constant related to the dimension of the matrices $A^{\kappa^\prime\hbar}$ and $A^{\kappa^\prime\hbar+\hbar}$. Since $\{|A_{ij}^{\kappa^\prime\hbar}|_{\max}\}_{\kappa^\prime=1}^K$ can be viewed as the enveloping stochastic process of $\{A_{ij}^{\kappa^\prime\hbar}\}_{\kappa^\prime=1}^K$, we conjecture that the weak dependence between them exhibit similarities.
Thus, $\{A^{\kappa^\prime\hbar}\}_{\kappa^\prime=1}^{K}$ satisfies the conditions of Lemma~\ref{lemma:main_weak}, ensuring that it is a weakly dependent subsequence constructed from $\{A^{\kappa^\prime}\}_{\kappa^\prime=1}^{K}$. Consequently, the sequence $\{\Delta^{\kappa^\prime\hbar}(\boldsymbol{x})\}_{\kappa^\prime=1}^{K}$ is weakly dependent. 

To sum up the conclusions above, the sequences $\{\boldsymbol{z}^{\kappa^\prime\hbar-1}(\boldsymbol{x})\}_{\kappa^\prime=1}^{K}$ and $\{\Delta^{\kappa^\prime\hbar}(\boldsymbol{x})\}_{\kappa^\prime=1}^{K}$ are both weakly dependent, ensuring the weak dependence of the sequence $\{\text{NTK}_{(d)}^{\kappa^\prime\hbar}\}_{\kappa^\prime = 1}^{K}$. From~\citep[chap.~1.3]{doukhan2012:mixing}, it is observed that a $\beta$-mixing sequence with an exponential rate of convergence is covered by an $\alpha$-mixing sequence with a rate of $\mathcal{O}(n^{-5})$. Thus, the sequence $\{\text{NTK}_{(d)}^{\kappa^\prime\hbar}\}_{\kappa^\prime = 1}^{K}$ satisfies the conditions of Lemma~\ref{weakCLT}, that is, the Generalized Central Limit Theorem. Therefore, NTK$_{(d)}$ converges to a Gaussian distribution as $K\rightarrow +\infty$, which completes the proof of Theorem~\ref{thm:NTK_depth}. \hfill$\blacksquare$

{\bf Discussions on Theorem~\ref{thm:NTK_depth}.} To the best of our knowledge, our proposed NTK$_{(d)}$ kernel is the first NTK kernel derived by increasing network depth. It is observed that the shortcut-related architecture plays a key role in producing the NTK$_{(d)}$ kernel. With this architecture, we ensure that the product of the weight gradients of the layers separated by $\hbar$ and the corresponding weight matrix derives a weakly dependent sequence. Thus, the NTK$_{(d)}$ kernel converges to a Gaussian distribution as the number of shortcut connections $K$ goes to infinity. Based on the analysis above, the characteristics of the NTK$_{(d)}$ kernel are closely related to the network depth, also known as the separation constant $\hbar$ and the number of shortcut connections $K$. Notice that in practice, $\hbar$ is not necessarily a constant; it suffices to enable a large $\hbar$ to ensure the weak dependence of the sequence $\{\text{NTK}_{(d)}^{\kappa^\prime\hbar}\}_{\kappa^\prime=1}^K$. While for $K$, it should approach infinity, which ensures the existence of the NTK$_{(d)}$ kernel.

\subsection{Spectrum of the Depth-induced NTK Kernel}\label{subsec: bound_it}

In this subsection, we investigate the spectrum of the depth-induced NTK kernel NTK$_{(d)}$, which is closely tied to the generalization and training stability of deep neural networks. Specifically, we establish a lower bound on its smallest eigenvalue and an upper bound on its largest singular value.
\begin{theorem}\label{thm:smallest_eigenvalue}(Lower Bound on the Smallest Eigenvalue) Let $\left\{\boldsymbol{x}_i\right\}_{i=1}^{N}$ be a set of i.i.d. data points from a data distribution $P_X$. Provided the shortcut-related network defined by Eqs.~\eqref{eq:forward} and~\eqref{eq:shortcut}, if the followings hold
	\begin{itemize}
		\item[(1)] the activation $\phi$ is Leaky ReLU, that is, $\phi(x) = max(x,\alpha x)$ where $\alpha \in (0,1)$,
		\item[(2)] the weight matrices are stable-pertinent for $\phi$, that is, $\mathbf{W}^l \in SP(\phi)$ for $\forall~ l\in [L]$,
		\item[(3)] the data distribution $P_X$ is well scaled,
	\end{itemize}
	then, we have for any even integer constant $r$ that satisfies $r \geq 2$, it holds with the probability of at least \(1-N\exp({-\Omega(d)})-N^2\exp({-\Omega(dN^{-2/(r-0.5)}))}\), that the smallest eigenvalue of the depth-induced NTK kernel is lower bounded by
	\[\lambda_{\min}(\text{NTK}_{(d)}) \geq \mu_r(\phi)^2\Omega(d)\ ,\]
	where $\mu_r(\phi)$ is the $r$-th Hermite coefficient of the Leaky ReLU function $\phi$.
\end{theorem}

Theorem~\ref{thm:smallest_eigenvalue} establishes a lower bound on the smallest eigenvalue of the depth-induced NTK kernel NTK$_{(d)}$, which is related to the input dimension $d$, and does not hide any other dependencies on the network depth $L$. It is obtained under well-scaled data assumptions, in line with conventional studies on the width-induced NTK kernel~\citep{simon2019gradient,pmlr-v139-nguyen21g}. According to Theorem~\ref{thm:smallest_eigenvalue}, the NTK$_{(d)}$ kernel cannot degenerate to a zero kernel. Moreover, it has been found by previous research that the lower bound of the smallest eigenvalue of NTK$_{(d)}$ is highly related to the generalization property of deep neural networks~\citep{arora2019fine,chen2020generalized,simon2019gradient}. Thus, our future work focuses on exploring how the smallest eigenvalue of NTK$_{(d)}$ affects the generalization performance, and whether tighter lower bounds can lead to practical benefits in the performance of deep neural networks. We provide a proof sketch of Theorem~\ref{thm:smallest_eigenvalue} below, and the full proof can be accessed in Appendix~\ref{appendix: full_proof_small}. Further, we validate the lower bound on the smallest eigenvalue of NTK$_{(d)}$ by numerical experiments in Subsection~\ref{subsec: lower_bound_exp}.

\noindent\textit{Proof Sketch.} For simplicity, we force $n = n_o =  n_\hbar = n_{2\hbar}  = \cdots = n_{K\hbar}$, while keeping $d = n_0$ independent from $n$. Thus, we can omit the all-ones matrices $\mathbf{J}^{\kappa\hbar}$ in Eq.~\eqref{eq:shortcut}. Also, we neglect the scaling constant $1/\sqrt{M_{\boldsymbol{z}}}$, which is barely related to the input dimension $d$.

The proof relies on the following inequalities about the smallest eigenvalues of matrices
\begin{equation}\label{real_main_lambda_min}
	\lambda_{\min}(A+B) \geq \lambda_{\min}(A) + \lambda_{\min}(B)\quad \text{and} \quad
	\lambda_{\min}(A\circ B) \geq \lambda_{\min}(A){\min}_{i\in[n]}B_{ii}\ ,
\end{equation}
where the first inequality holds for Hermitian matrices~\citep{weyl1949inequalities} and the second one holds for positive semi-definite matrices~\citep{Schur1911}. By combining Eqs.~\eqref{main_NTK_sum} and~\eqref{real_main_lambda_min}, it suffices to lower bound $\lambda_{\min}(\text{NTK}_{(d)}^{K\hbar})$. Let ${\mathbf{Z}^{K\hbar}}^\top$ denote the column vector vec$(\boldsymbol{z}^{K\hbar}(\boldsymbol{x}_i)^\top)_{i=1}^N$. Let ${\boldsymbol{\Delta}^{K\hbar}}^\top$ denote the column vector vec$(\Delta^{K\hbar}(\boldsymbol{x}_i)^\top)_{i=1}^N$. Then, we can prove that NTK$_{(d)}^{K\hbar}$ can be written as the form of a Hadamard product by
\begin{equation}\label{real_main_NTK_element}
	\text{NTK}_{(d)}^{K\hbar} = \left({\mathbf{Z}^{K\hbar}}^\top {\mathbf{Z}^{K\hbar}}\right) \circ \left({\boldsymbol\Delta^{K\hbar}}^\top{\boldsymbol\Delta^{K\hbar}}\right)\ .
\end{equation}
Invoking Eq.~\eqref{real_main_lambda_min} into Eq.~\eqref{real_main_NTK_element}, we have
\begin{equation}\label{real_main_apart_z_delta}
	\lambda_{\min}\left(\text{NTK}_{(d)}^{K\hbar}\right) \geq \lambda_{\min}\left({\mathbf{Z}^{K\hbar}}^\top {\mathbf{Z}^{K\hbar}}\right)\min_{i\in[n\times N]}\left({\boldsymbol\Delta^{K\hbar}}^\top{\boldsymbol\Delta^{K\hbar}}\right)_{ii}\ .
\end{equation}
By standard calculations, we can prove that the term of $\boldsymbol\Delta$ in Eq.~\eqref{real_main_apart_z_delta} is unrelated to the input dimension $d$ and cannot reduce to zero. Then, for the term of $\mathbf{Z}$, by using Hermite expansion of the Leaky ReLU, we can bound $\lambda_{\min}({\mathbf{Z}^{K\hbar}}^\top {\mathbf{Z}^{K\hbar}})$ in terms of $\lambda_{\min}(({\mathbf{Z}^{K\hbar-1}}^{(r)})^\top({\mathbf{Z}^{K\hbar-1}}^{(r)}))$, where $(\cdot)^{(r)}$ denotes the $r$-th Khatri Rao power. By recursively applying this argument, it suffices to bound $\lambda_{\min}(({\mathbf{X}}^{(r)})^\top({\mathbf{X}}^{(r)}))$, where ${\mathbf{X}}^\top$ denotes the column vector vec$(\boldsymbol{x}_i^\top)_{i=1}^N$. This can be done by the Gershgorin circle theorem~\citep{SALAS199915}, and by the well-scaled property of the data distribution $P_X$.
\hfill$\blacksquare$
\begin{theorem}\label{thm:largest_eigenvalue}(Upper Bound on the Largest Singular Value) Provided the shortcut-related network defined by Eqs.~\eqref{eq:forward} and~\eqref{eq:shortcut}, if the activation $\phi$ is well-posed and satisfy $|\phi(\cdot)| \leq 1$, and the weight matrices are stable-pertinent for $\phi$, then the largest singular value of $\text{NTK}_{(d)}$ is upper bounded by
	\[
	\sigma_{\max}(\text{NTK}_{(d)}(\boldsymbol{x},\boldsymbol{x}^\prime)) \leq \frac{K(2K+1)}{6}n_{\max}^2 \ ,
	\] 
	where $n_{\max}$ is the largest width of the concerned network. 
\end{theorem}
Theorem~\ref{thm:largest_eigenvalue} establishes the upper bound on the largest singular value of NTK$_{(d)}$, providing a fundamental insight for NTK$_{(d)}$ and neural kernel theory. The upper bound $\sigma_{\max}(\text{NTK}_{(d)}(\boldsymbol{x},\boldsymbol{x}^\prime)) = \mathcal{O}(K^2)$ indicates that the largest singular value of NTK$_{(d)}$ grows at most quadratically with $K$, leveraging the relation between the neural kernel spectrum and the number of shortcut connections, where the former is closely related to kernel regression and the latter indicates the network depth. This upper bound contributes to deriving the condition number of the NTK matrix between any two inputs, ensuring its stability. Besides, a smaller condition number of the NTK matrix is known to improve convergence and enhance training stability of the concerned neural network~\citep{liu2023relu}. The restraints on activations are mild and can be satisfied by commonly used ones, such as Sigmoid and Tanh. We provide the proof of Theorem~\ref{thm:largest_eigenvalue} below, and validate the upper bound on the largest singular value of NTK$_{(d)}$ by numerical experiments in Subsection~\ref{subsec: lower_bound_exp}.

\noindent\textit{Proof.} For simplicity, we force $n_{\max} = n_o = n_0 = n_\hbar = \cdots = n_{K\hbar}$. By the expanded form of NTK$_{(d)}$ in Theorem~\ref{thm:NTK_depth}, it is composed of two terms, the inner product of $\Delta$ and that of $\boldsymbol{z}$. We successively provide upper bounds for the largest singular value of both of them. We first give the upper bound of the term of $\Delta$, where
	\begin{equation}\label{delta_expression}
		\Delta^{\kappa^\prime\hbar}(\boldsymbol{x}) = \sum_{\kappa = \kappa^\prime}^{K}\left[\prod_{i=\kappa^\prime\hbar}^{\kappa\hbar}\left(\hat{\mathbf{W}}^{i}\right)^\top\mathbf{D}^{i}(\boldsymbol{x})\right]~{\mathbf{J}^{\kappa\hbar}}^\top\ .
	\end{equation}
	For two matrices $\mathbf{A}$ and $\mathbf{B}$, the following basic properties hold
	\begin{equation}\label{main_lambda}
		\sigma_{\max}(\mathbf{AB}) \leq \sigma_{\max}(\mathbf{A}) \sigma_{\max}(\mathbf{B})
		\quad\text{and}\quad
		\sigma_{\max}(\mathbf{A}+\mathbf{B}) \leq \sigma_{\max}(\mathbf{A})+\sigma_{\max}(\mathbf{B}) \ .
	\end{equation}
	Since the activation is well-posed and the weights are stable-pertinent, according to Eqs.~\eqref{delta_expression} and~\eqref{main_lambda}, we have
	\begin{equation} \label{main:lambda_kappa_prime}
		\sigma_{\max}(\Delta^{\kappa^\prime\hbar}(\boldsymbol{x})) \leq \sum_{\kappa = \kappa^\prime}^{K}\epsilon^{(\kappa-\kappa^\prime)\hbar+1} \sqrt{n_o ~n_{\kappa\hbar}}\leq (K+1-\kappa^\prime)n_{\max}\ .
	\end{equation}
	By combining Eqs.~\eqref{main_lambda} and~\eqref{main:lambda_kappa_prime}, we have
	\begin{equation}\label{lambda_delta}
		\sigma_{\max}(\langle\Delta^{\kappa^\prime\hbar}(\boldsymbol{x}),\Delta^{\kappa^\prime\hbar}(\boldsymbol{x}^\prime)\rangle) \leq {(K + 1 -\kappa^\prime)}^2n_{\max}^2\ .
	\end{equation}
	For the term of $\boldsymbol{z}$, since the activation $\phi$ satisfies $|\phi(\cdot)| \leq 1$, $\boldsymbol{z}^{\kappa^\prime\hbar-1}(\boldsymbol{x})$ is a vector with elements less than or equal to $1$. Consequently, by Eq.~\eqref{main_lambda}, we obtain
	\begin{equation}\label{lambda_z}
		\sigma_{\max}(\langle\boldsymbol{z}^{\kappa^\prime\hbar-1}(\boldsymbol{x}),\boldsymbol{z}^{\kappa^\prime\hbar-1}(\boldsymbol{x}^\prime)\rangle)\leq n_{\max}\ .
	\end{equation}
	Finally, by substituting Eqs.~\eqref{lambda_delta} and~\eqref{lambda_z} into Eqs.~\eqref{main_NTK_sum} and~\eqref{main_NTK_exp}, and also by Eq.~\eqref{main_lambda}, we can further give the upper bound of $\sigma_{\max}(\text{NTK}_{(d)}(\boldsymbol{x},\boldsymbol{x}^\prime))$ as follows
	\begin{equation*}\sigma_{\max}(\text{NTK}_{(d)}(\boldsymbol{x},\boldsymbol{x}^\prime)) \leq \frac{ n_{\max} }{{M_{\boldsymbol{z}}}} ~ \sum_{\kappa^\prime = 1}^{K}(K + 1 -\kappa^\prime)^2n_{\max}^2 
		\leq \frac{K(2K+1)}{6}n_{\max}^2\ ,
	\end{equation*} 
	which completes the proof. $\hfill\blacksquare$

{\bf Discussions on Spectrum of Depth-induced NTK.} The spectrum of the NTK$_{(d)}$ kernel highlight two important aspects. (1) In line with conventional studies on the NTK$_{(w)}$ kernel~\citep{simon2019gradient,pmlr-v139-nguyen21g}, we establish a lower bound on the smallest eigenvalue of the NTK$_{(d)}$ kernel by Theorem~\ref{thm:smallest_eigenvalue}, which characterizes the generalization performance of deep neural networks~\citep{arora2019fine,chen2020generalized,simon2019gradient}. (2) Besides, unlike the NTK$_{(w)}$ kernel, which arises from shallow neural networks, the NTK$_{(d)}$ kernel is derived from infinitely deep neural networks. This necessitates additional analysis to ensure that the kernel is stable and does not explode as the depth increases. Thus, we further establish an upper bound on its singular values by Theorem~\ref{thm:largest_eigenvalue}, which explicitly captures the dependence of NTK$_{(d)}$ on the number of shortcut connections $K$ that is directly linked to the network depth $L$. Notice that singular values are adopted instead of eigenvalues when establishing the upper bound of NTK$_{(d)}$, as they offer a more convenient characterization under our analytical framework.

\subsection{Invariance during Training} \label{subsec: theory_invariance}
In this subsection, we show that our proposed $\text{NTK}_{(d)}$ kernel remains invariant during training, as well as the conventional width-induced NTK kernel~\citep{WideNTK2021}. For simplicity, we force $n = n_o = n_0 = n_\hbar = \cdots = n_{K\hbar}$. Let $\mathcal{D}$ denote the distribution on the input space $\mathbb{R}^{d}$, and $\mathcal{F}$ denotes the function space related to the shortcut-related network. On this space, we also consider the semi-norm $\| \cdot \|_{\mathcal{D}}$ for two functions $g,h: \mathbb{R}^{n_0} \rightarrow \mathbb{R}^{n_o}$ like Jacot et al.~\citep{WideNTK2021}, which is defined by $\langle g, h \rangle_{\mathcal{D}} = \mathbb{E}_{\boldsymbol{x} \sim \mathcal{D}} [ g(\boldsymbol{x})^\top h(\boldsymbol{x}) ]$. Given the initialized parameters \( \mathbf{W}^l \) for $l\in[L]$ and a training direction \( t \mapsto d_t \in \mathcal{F} \), the evolution of the network follows the differential equation as follows
\begin{equation}\label{eq: dt}
	\frac{\partial \mathbf{W}^l}{\partial t} = \left\langle \frac{\partial f}{\partial \mathbf{W}^l}, d_t \right\rangle_{\mathcal{D}}.  
\end{equation}

Based on the introduction above, we present our theorem as follows.
\begin{theorem}\label{thm:constant} (Training Invariance of Depth-induced NTK) Provided the shortcut-related network defined by Eqs.~\eqref{eq:forward} and~\eqref{eq:shortcut}, if the following conditions hold
	\begin{itemize}
		\item[(1)] $\phi$ is well-posed and satisfies the Lipschitz condition with $\|\phi(\boldsymbol{x}) - \phi(\boldsymbol{y})\|_s \leq L\|\boldsymbol{x}-\boldsymbol{y}\|_s$,
		\item[(2)] the weight matrices are stable-pertinent for $\phi$, that is, $\mathbf{W}^l \in SP(\phi)$ for $\forall~ l\in [L]$,
		\item[(3)] we consider any T such that the integral $\int_0^T\|d_t\|_{\mathcal{D}}\, dt$ remains stochastically bounded,
	\end{itemize}
	then, given a constant $\lambda$ satisfying $0 < \lambda<1$, as $K\rightarrow+\infty$ and $n\rightarrow+\infty$ with the ratio of $n=\Omega(K^{2/\lambda})$, it holds uniformly that $\text{NTK}_{(d)}(t) \rightarrow \text{NTK}_{(d)}(0)$ for $t\in[0,T]$.
\end{theorem}
Theorem~\ref{thm:constant} states that our proposed NTK$_{(d)}$ kernel remains invariant during training, enabling a precise description of the network's training dynamics by kernel regression, and extending the applicability of Theorems~\ref{thm:NTK_depth},~\ref{thm:smallest_eigenvalue} and~\ref{thm:largest_eigenvalue} to the entire training process. Besides, there is a clear limiting relation between $n$ and $K$, in which the width $n$ as well as the number of shortcut connections $K$ goes to infinity with a lower bound of $n=\Omega(K^{2/\lambda})$ where $0 < \lambda < 1$. An infinitely large $K$ is necessary for the existence of NTK$_{(d)}$ by Theorem~\ref{thm:NTK_depth}, while an infinitely large $n$ is employed to ensure that the change of NTK$_{(d)}$ is small enough during training. The above settings are consistent with those of Jacot et al.~\citep{WideNTK2021}, and we validate the training invariance of NTK$_{(d)}$ by numerical experiments in Subsection~\ref{subsec: exp_invariance}. The full proof of Theorem~\ref{thm:constant} can be accessed in Appendix~\ref{proof_constant}.
\vspace{-10pt}

\section{Experiments}\label{Experiments}
In this section, we conduct numerical experiments to evaluate the effectiveness of the proposed depth-induced NTK kernel NTK$_{(d)}$, comparing it to the traditional width-induced NTK kernel NTK$_{(w)}$ and verifying the properties of the NTK$_{(d)}$ kernel. The experiments are performed to discuss the following questions:
\begin{enumerate}
	\item Is the performance of the proposed NTK$_{(d)}$ kernel comparable with that of the traditional NTK$_{(w)}$ kernel?
	\item Whether and to what extent do the separation constant $\hbar$ and the number of shortcut connections $K$ affect the characteristics of the NTK$_{(d)}$ kernel? 
	\item How do the spectrum of NTK$_{(d)}$ manifest in practice, particularly its smallest eigenvalue and largest singular value?
	\item Does the NTK$_{(d)}$ kernel remain invariant during training as Theorem~\ref{thm:constant} states?
\end{enumerate}

\begin{table}[t]
	\centering
	\vskip 0.15in
	
	\begin{tabular}{lcccccr}
		\toprule
		\textbf{Exp.} & \textbf{Purpose} & \textbf{Network} & \textbf{Width} & $K$ & $\hbar$  \\
		\midrule
		\multirow{2}{*}{~\ref{subsec:Kernel regression on sine function}}  &\multirow{2}{*}{Basic Performance} & Wide  & 500 & -- & --  \\
		& &  Deep & 20  & 25 & 2  \\
		\midrule
		\multirow{2}{*}{~\ref{subsec:Kernel Regression on Fashion-MNIST}} & \multirow{2}{*}{Basic Performance} & Wide  & 200 & -- & --  \\
		&  & Deep & 20  & 25 & 2  \\
		\midrule
		{~\ref{subsec:Effects of interval hbar and depth K on NTK(d)}}& Parameter Influence & Deep  & 20 & \{10,30,50\} & \{1,2,3\}  \\
		\midrule
		\multirow{2}{*}{~\ref{subsec: lower_bound_exp}}  & Smallest Eigenvalue & Deep  & 20 & 25 & 2  \\
		&Largest Singular Value &  Deep & 20  & \{10,20,$\cdots$,100\} & 2  \\
		\midrule
		{~\ref{subsec: exp_invariance}}& Training Invariance & Deep  & \{25,100,225\} & \{5,10,15\} & 3  \\
		\bottomrule
	\end{tabular}
	\caption{Experimental configurations.}
	\label{configurations of experiments}
\end{table}

{\bf Configurations.} Our experiments follow the main configurations of Lee et al.~\citep{lee2018:NNGP}. For implementing NTK$_{(w)}$, we employ a single-hidden-layer network architecture with a large width, referred to as the wide architecture. By contrast, for NTK$_{(d)}$, we adopt the shortcut-related architecture described in Subsection~\ref{subsec: Topology of Neural Network}, where all hidden layers have the same width, referred to as the deep architecture. Here, we neglect the scaling factor $1/\sqrt{n_l}$ in Eq.~\eqref{eq:forward} for simplicity. Both networks employ ReLU as the activation and are trained by vanilla gradient descent. All the weights $\mathbf{W}$ and biases $\mathbf{b}$ are independently initialized with centered Gaussian distributions. We set the learning rate to $0.001$ and the batch size to $64$. Table~\ref{configurations of experiments} displays the purpose of each experiment, the settings of network width, separation constant $\hbar$, and the number of shortcut connections $K$ in our experiments. For each experiment, we repeat $10$ trials for statistical rigor. All experiments are conducted on the NVIDIA GeForce RTX 3060 Laptop GPU. 

{\bf Kernel Regression.} Our experiments employ the method of kernel regression. Let $\mathbf{X} = \{\boldsymbol{x}_1,\boldsymbol{x}_2,\cdots,\boldsymbol{x}_n\}$ be the set of training samples where $\boldsymbol{x}_i\in\mathbb{R}^d$ for $i\in [n]$. Let $\mathbf{Y} = \{\boldsymbol{y}_1,\boldsymbol{y}_2,\cdots,\boldsymbol{y}_n\}$ be the set of corresponding labels where \(\boldsymbol{y}_i\in\mathbb{R}^o\) for $i\in [n]$. Provided the set of testing samples $\mathbf{X}^*$, the goal is to predict their labels by learning $f:\mathbf{X}\cup\mathbf{X}^* \rightarrow \mathbf{Y}\cup\mathbf{Y}^*$. Let $K(\boldsymbol{x}_i,\boldsymbol{x}_j):\mathbb{R}^{d}\times\mathbb{R}^{d}\rightarrow\mathbb{R}$ be a kernel for any inputs $\boldsymbol{x}_i,\boldsymbol{x}_j$ where $i,j\in [n]$. Provided that $f(\mathbf{X})$ and $f(\mathbf{X}^*)$ belong to a multivariate Gaussian distribution $\mathcal{N}(0,\mathbf{\Sigma})$, where
\begin{equation*} \label{eq:kernel}
	\mathbf{\Sigma} = 
	\begin{bmatrix}
		K({\mathbf{X},\mathbf{X}}) &  K({\mathbf{X}^*, \mathbf{X})}^{\top}\\
		K{(\mathbf{X}^*, \mathbf{X})} &  K({\mathbf{X}^*,\mathbf{X}^*})  
	\end{bmatrix} \ ,
\end{equation*}
the predictions of kernel regression are produced by
\begin{equation*} 
	\left\{ \begin{aligned}
		& \mu^* = K({\mathbf{X}^*, \mathbf{X})} {K({\mathbf{X},\mathbf{X}})}^{-1} \boldsymbol{Y}^{\top} \ , \\
		& \mathbf{\Sigma}^* = K({\mathbf{X}^*,\mathbf{X}^*}) - K({\mathbf{X}^*, \mathbf{X}}) {K(\mathbf{X},\mathbf{X})}^{-1} {K(\mathbf{X}^*, \mathbf{X})}^{\top} \ , 
	\end{aligned} \right.
\end{equation*}
where $\mu^*$ and $\mathbf{\Sigma}^*$ denote the mean and covariance of regression predictions, respectively. In our experiments, we employ the traditional width-induced NTK$_{(w)}$ and the proposed depth-induced NTK$_{(d)}$ as kernels separately. To facilitate the implementation of regression experiments, we employ the trace of the NTK kernel matrix for any two samples instead of the entire matrix.

To gain a deeper understanding of the characteristics of a kernel, we investigate the similarities and differences between samples in the kernel space. For any two samples $\boldsymbol{x}, {\boldsymbol{x}}^\prime$, we define the angle $\alpha(\boldsymbol{x}, {\boldsymbol{x}}^\prime)$ between them in the kernel space by
\[
\alpha(\boldsymbol{x},{\boldsymbol{x}}^\prime) = \arccos\left(\frac{K(\boldsymbol{x},{\boldsymbol{x}}^\prime)}{\sqrt{K(\boldsymbol{x},\boldsymbol{x})~ K(\boldsymbol{x}^\prime,{\boldsymbol{x}}^\prime)}}\right).
\]
Under our experimental settings, the angles range from $[0,\pi/2]$, where values close to $0$ indicate that the corresponding samples are highly similar in the kernel space, while values near $\pi/2$ suggest significant differences. The distribution of the angles reflects the kernel's ability to differentiate between samples. If most angles cluster near $0$, it implies that the kernel fails to distinguish between different samples, making it less effective.

\begin{figure}[t]
	\centering
	\includegraphics[width=0.58\linewidth]{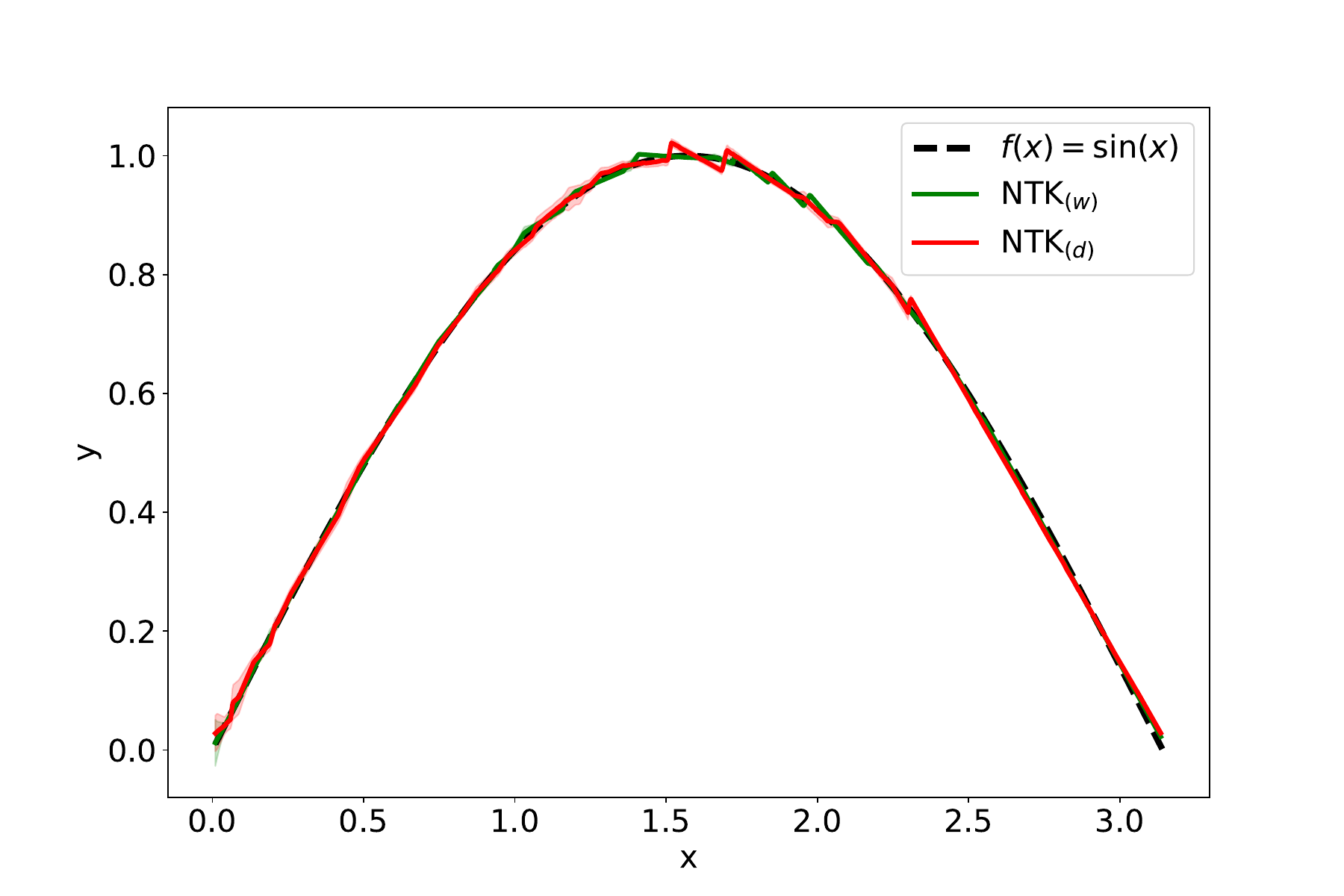}
	\vspace{-10pt}
	\caption{Fitting curves of kernel regression with NTK$_{(w)}$ and NTK$_{(d)}$ on sine function.}
	\label{fig:sinfunction}
\end{figure}

\subsection{Kernel Regression on Sine Function}\label{subsec:Kernel regression on sine function}

\begin{figure}[p]
	\centering
	\includegraphics[width=0.98\linewidth]{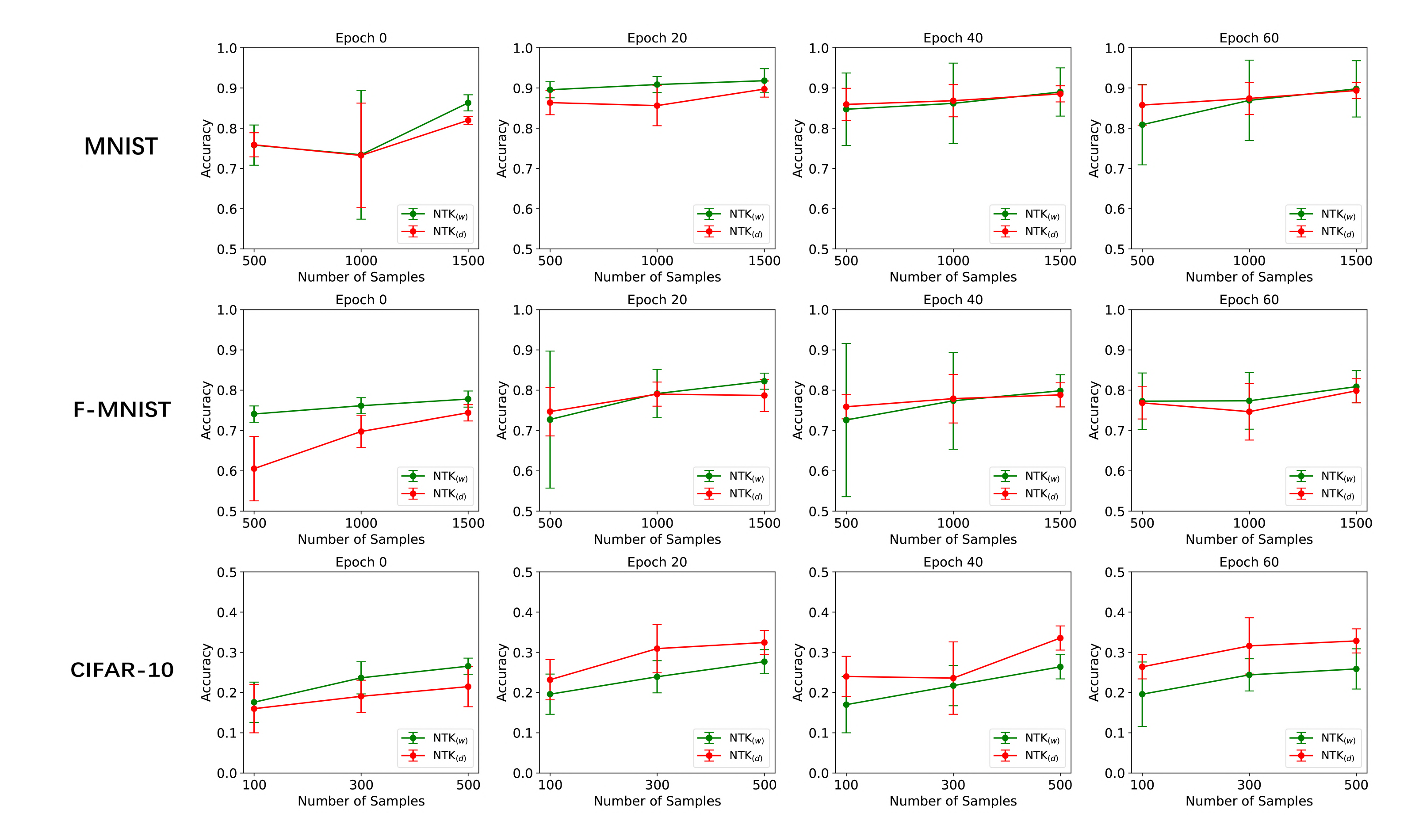}
	\caption{The error bar plots of the regression accuracy of NTK$_{(w)}$ and NTK$_{(d)}$ on MNIST, Fashion-MNIST, and CIFAR-10 data sets.}
	\label{fig: combined_error_bars}
\end{figure}

\begin{figure}[p]
	\centering
	\includegraphics[width=0.98\linewidth]{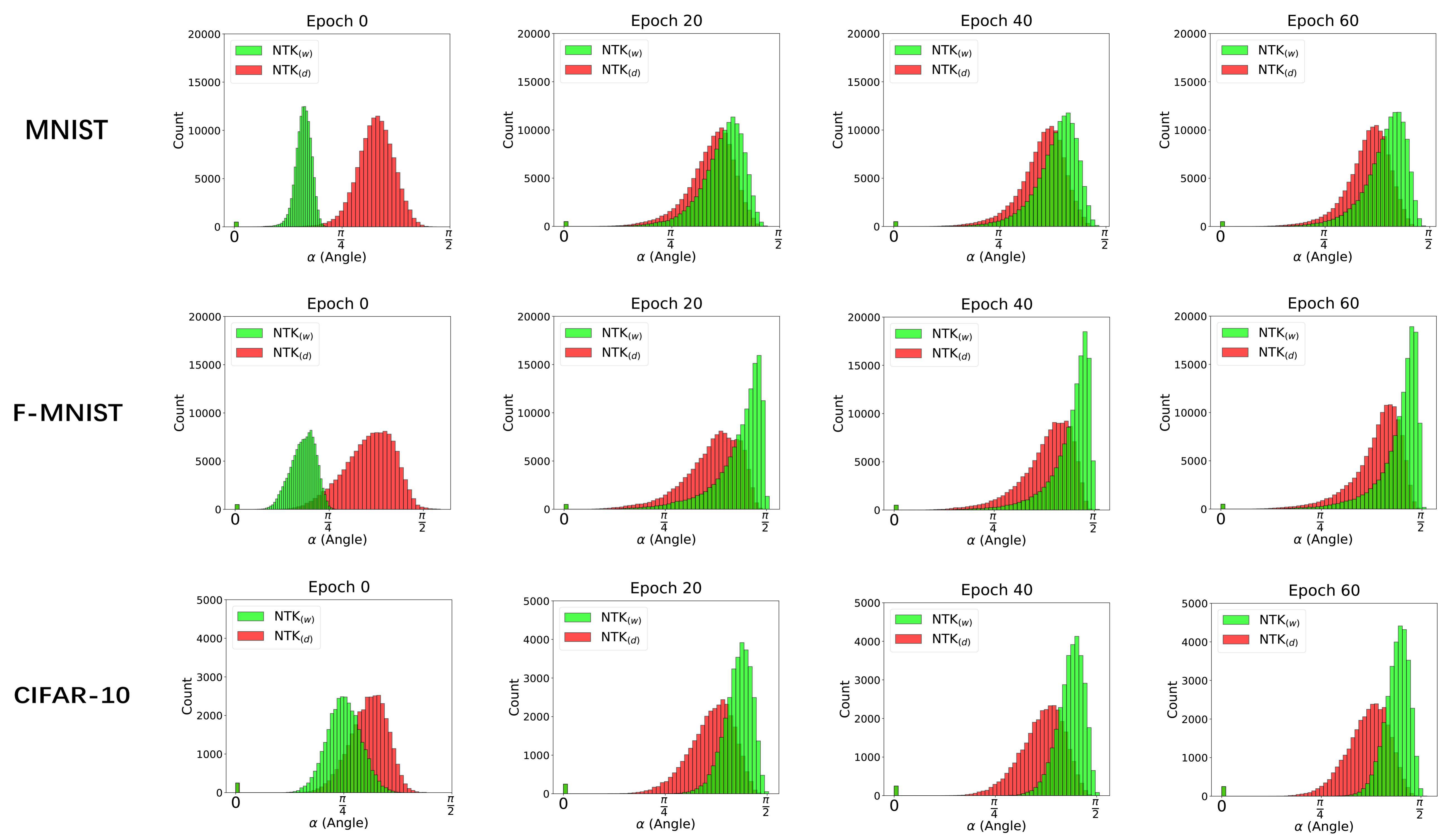}
	\caption{The angular plots of NTK$_{(w)}$ and NTK$_{(d)}$ on MNIST, Fashion-MNIST, and CIFAR-10 data sets.}
	\label{fig: NTK_width_depth_comparison}
\end{figure}

Following the experimental configurations of Zhang et al.~\citep{Zhang2024:deep}, we first employ the NTK$_{(w)}$ and NTK$_{(d)}$ kernels to fit a function $f(x) = \sin(x)$ over $[0,\pi]$. A total of $500$ data points, divided into a training set of 300 points and a testing set of the remaining 200 points, are randomly sampled across the interval $[0, \pi]$. 

The NTK$_{(w)}$ and NTK$_{(d)}$ kernels are gained from training two networks with wide and deep architectures for $5$ epochs, respectively. The weight matrices $\mathbf{W}^l$ and the biases $\boldsymbol{b}^l$ of the wide and deep neural networks are independently initialized with Gaussian distributions for $l\in[L]$, that is, $\mathbf{W}^l\sim\mathcal{N}(0,0.02)$ and $\boldsymbol{b}^l\sim\mathcal{N}(0,0.01)$ for the wide network, and $\mathbf{W}^l\sim\mathcal{N}(0,0.2)$ and $\boldsymbol{b}^l\sim\mathcal{N}(0,0.1)$ for the deep one. Table~\ref{configurations of experiments} lists the detailed configurations of this experiment.

Figure~\ref{fig:sinfunction} shows the fitting curves of kernel regression using NTK$_{(w)}$ and NTK$_{(d)}$ within a 95\% confidence interval. It is observed that both regression curves fit the sine function well. The fluctuations that appear on the interval $[1.5,2]$ may be caused by fewer samples. From this observation, we conclude that the performance of NTK$_{(d)}$ is comparable to NTK$_{(w)}$ in the task of fitting the sine function.

\subsection{Kernel Regression on MNIST, Fashion-MNIST, and CIFAR-10}\label{subsec:Kernel Regression on Fashion-MNIST}
To further demonstrate the effectiveness of NTK$_{(d)}$, we extend kernel regression experiments on the \href{https://www.kaggle.com/datasets/hojjatk/mnist-dataset}{MNIST}, \href{https://www.kaggle.com/datasets/zalando-research/fashionmnist}{Fashion-MNIST}, and \href{https://www.kaggle.com/c/cifar-10/}{CIFAR-10} data sets, following the main configurations of Lee et al.~\citep{lee2018:NNGP}. For the MNIST and Fashion-MNIST data sets, we separately sample $\{500, 1000, 1500\}$ data points, dividing them equally between the training and testing sets. For the CIFAR-10 data set, the input images are larger with three color channels, increasing the computational complexity. Thus, we slightly reduce the number of samples for regression to $\{100,300,500\}$, also dividing them equally between the training and testing sets.

The NTK$_{(w)}$ and NTK$_{(d)}$ kernels are gained from training two networks with wide and deep architectures for $60$ epochs, respectively. Here, we adopt the configurations listed in Table~\ref{configurations of experiments} to ensure a similar amount of network parameters concerning NTK$_{(w)}$ and NTK$_{(d)}$. We also limit the network width of the deep architecture to only $20$ so that the power of NTK$_{(d)}$ mainly comes from the depth. All the weights $\mathbf{W}$ and biases $\mathbf{b}$ are independently initialized by $\mathbf{W}^{l}\sim\mathcal{N}(0,0.4/n_l)$ and $\mathbf{b}^{l}\sim\mathcal{N}(0,0.2/n_l)$ for $l\in[L]$.

Figure~\ref{fig: combined_error_bars} shows the error bar plots of the regression accuracy of NTK$_{(w)}$ and NTK$_{(d)}$ kernels on different numbers of samples on the MNIST, Fashion-MNIST, and CIFAR-10 data sets across the training epochs of the networks. For each data set, it is observed that at initialization, the regression accuracy of NTK$_{(d)}$ is lower than NTK$_{(w)}$. However, when the networks are trained, the regression accuracy of NTK$_{(d)}$ becomes similar to NTK$_{(w)}$ on the MNIST and Fashion-MNIST data sets, and significantly surpasses it on the CIFAR-10 data set. From these observations, we can conclude that NTK$_{(d)}$ has a comparable ability to NTK$_{(w)}$. Additionally, we notice that when the networks are trained, the standard deviation of the regression accuracy of NTK$_{(d)}$ is usually smaller than NTK$_{(w)}$, suggesting that NTK$_{(d)}$ yields more stable regression performance.

Figure~\ref{fig: NTK_width_depth_comparison} presents the angular plots of the NTK$_{(w)}$ and NTK$_{(d)}$ kernels on the MNIST, Fashion-MNIST, and CIFAR-10 data sets, using $500$ training samples from MNIST and Fashion-MNIST, and $300$ from CIFAR-10. The results are consistent across all three data sets. It is observed that at initialization, the angles between samples in the kernel space of NTK$_{(w)}$ predominantly lie within $[0,\pi/4]$, whereas for NTK$_{(d)}$, they are mainly distributed within $[\pi/4,\pi/2]$. This suggests that at initialization, NTK$_{(d)}$ emphasizes capturing the differences between samples, whereas NTK$_{(w)}$ primarily focuses on their similarities. As the training of the networks progresses, the plots of NTK$_{(w)}$ and NTK$_{(d)}$ both move rightward, indicating that both kernels have improved their ability to distinguish between different samples. Throughout the training processes, the change of the plot of NTK$_{(d)}$ is smaller than that of NTK$_{(w)}$, indicating better stability of the NTK$_{(d)}$ kernel during training. 

\begin{figure}[p]
	\centering
	\includegraphics[width=\linewidth]{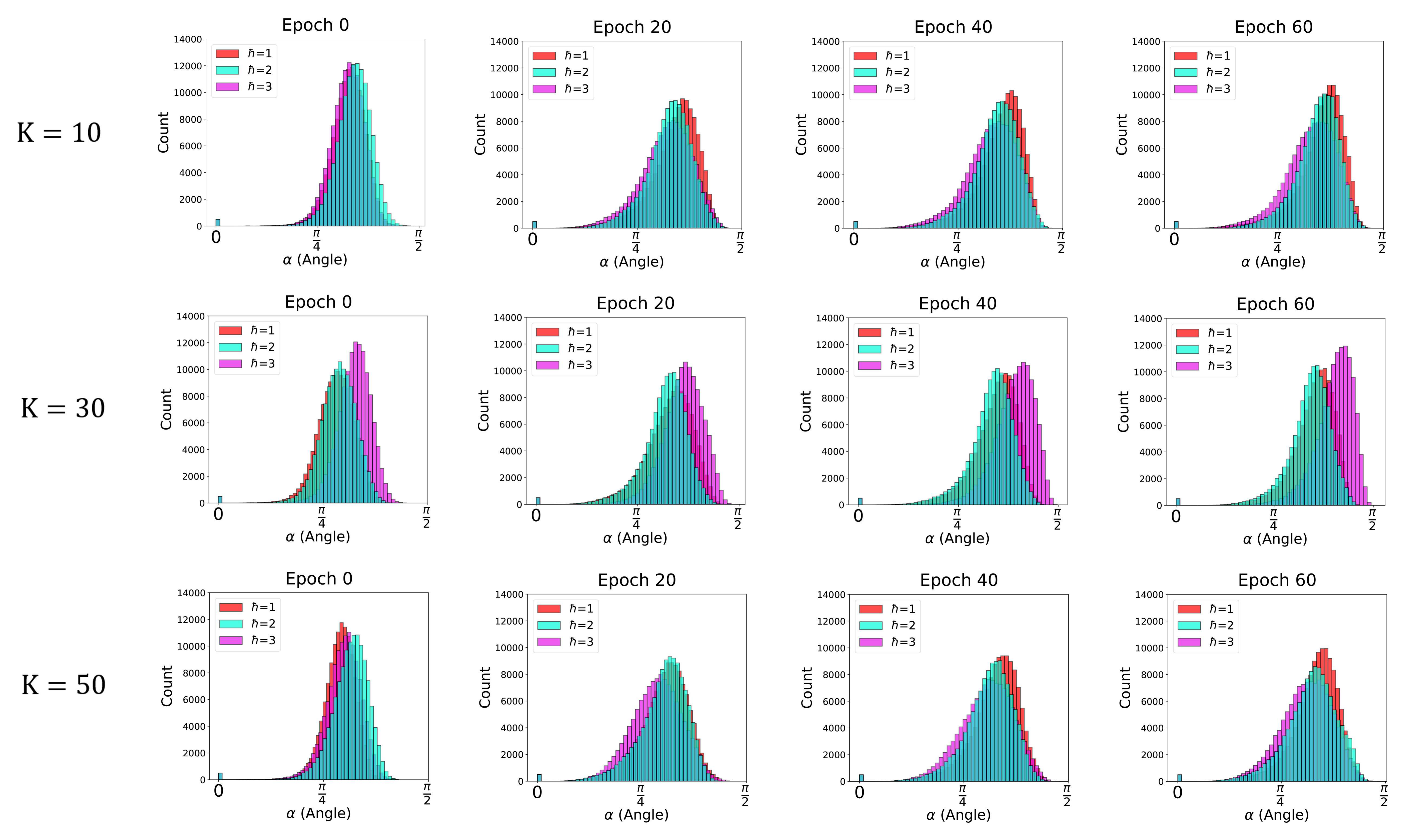}
	\vspace{-8pt}
	\caption{The angular plots of NTK$_{(d)}$  with different $\hbar$.}
	\label{fig:different_hbar}
\end{figure}

\begin{figure}[p]
	\centering
	\includegraphics[width=\linewidth]{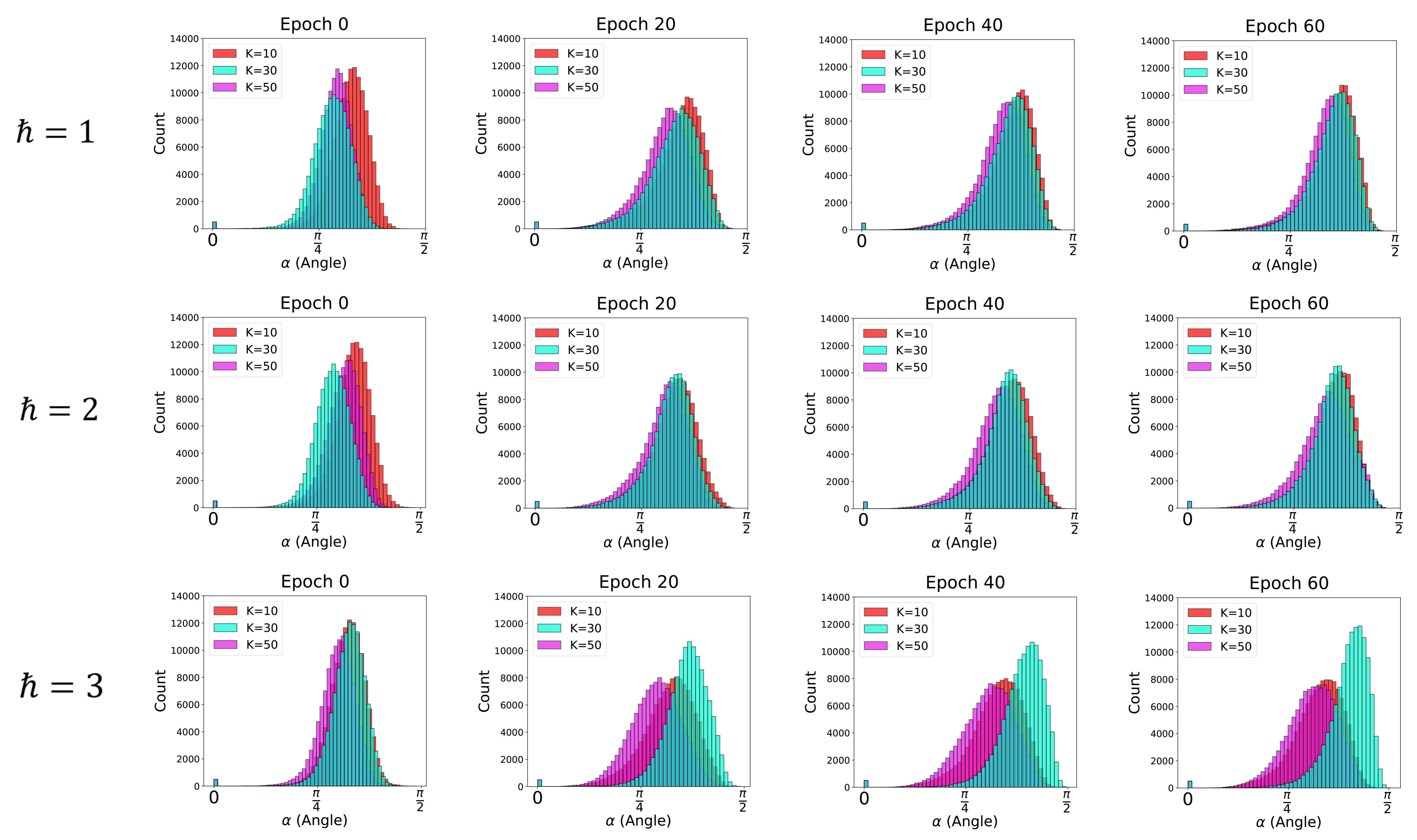}
	\vspace{-8pt}
	\caption{The angular plots of NTK$_{(d)}$  with different $K$.}
	\label{fig:different_K}
\end{figure}

\subsection{Effects of \texorpdfstring{$\hbar$}{hbar} and \texorpdfstring{$K$}{K}}\label{subsec:Effects of interval hbar and depth K on NTK(d)}

From Theorem~\ref{thm:NTK_depth}, the characteristics of the NTK$_{(d)}$ kernel are closely related to the network depth, depicted by the separation constant $\hbar$ and the number of shortcut connections $K$. In this subsection, we separately investigate the effect of $\hbar$ and that of $K$ on the characteristics of the proposed NTK$_{(d)}$ kernel, following the main configurations of Lee et al.~\citep{lee2018:NNGP}. The detailed experimental configurations are listed in Table~\ref{configurations of experiments}. For convenience, we here conduct experiments on the MNIST data set and take a total of $500$ training and $500$ testing samples. All the weights $\mathbf{W}$ and biases $\mathbf{b}$ of the concerning neural networks are independently initialized by $\mathbf{W}^{l}\sim\mathcal{N}(0,0.4/n_l)$ and $\mathbf{b}^{l}\sim\mathcal{N}(0,0.2/n_l)$ for $l\in[L]$.

We first investigate the effect of $\hbar$ by fixing $K$. Figure~\ref{fig:different_hbar} shows the angular plots of NTK$_{(d)}$ with different $\hbar$ on the MNIST data set across the training epochs, where $K$ is separately fixed as $10,30,50$ in each row. It is observed that there is a significant overlap between the plots of NTK$_{(d)}$ with $\hbar = 1,2,3$ across the training epochs for all the subfigures. From this observation, we conjecture that the separation constant $\hbar$ has little influence on the characteristics of NTK$_{(d)}$. 

We then investigate the effects of $K$ by fixing $\hbar$. Figure~\ref{fig:different_K} shows the angular plots of NTK$_{(d)}$ with different $K$ on the MNIST data set across the training epochs, where $\hbar$ is separately fixed as $1,2,3$ in each row. There are two significant findings: (1) For each subfigure of NTK$_{(d)}$ with $K = 10,30,50$, the distributions of concerned angles at initialization are different from those at the training phase. As training progresses, the distributions of the angles gradually shift toward larger values, implying that the corresponding kernels have improved their ability to distinguish between different samples. (2) Some subfigures of NTK$_{(d)}$ with $K = 10,30,50$ differ from each other. Especially in the case of $\hbar=3$ and $t=20,40,60$, the plot that corresponds to $K=30$ is positioned farthest to the right in comparison with those of $K=10$ and $K=50$, implying that NTK$_{(d)}$ with $K=30$ has a better ability to capture the differences between different samples. Therefore, we conclude that the number of shortcut connections $K$ greatly influences the characteristics of NTK$_{(d)}$.

From the observations above, the number of shortcut connections $K$ of the concerned neural network has a more significant impact on the characteristics of the NTK$_{(d)}$ kernel than the separation constant $\hbar$. This empirical finding is consistent with our theoretical results of Theorem~\ref{thm:NTK_depth}, which state that the proposed NTK$_{(d)}$ kernel highly relies on $K$ approaching infinity, while it only requires a large enough $\hbar$ to ensure the weak dependence of the sequence $\{\text{NTK}_{(d)}^{\kappa^\prime\hbar}\}_{\kappa^\prime=1}^K$. 

\subsection{Spectrum of the Depth-induced NTK Kernel}\label{subsec: lower_bound_exp}
We first investigate the smallest eigenvalue of the NTK$_{(d)}$ kernel by following the experimental configurations of Oymak and Soltanolkotabi~\citep{oymak2020toward} and Nguyen et al.~\citep{pmlr-v139-nguyen21g}. The data set is simulated, with $d$-dimensional input on the unit circle and single-dimensional output, and the labels are standard normal variables. We separately pick $N = 250, 500, 750$ as the number of samples for both training the shortcut-related deep neural networks and computing the NTK$_{(d)}$ kernel. The input dimension $d$ is varied across $0, 100, 200, \dots, 500$. For each combination of $N$ and $d$, we perform $10$ independent runs of training the concerned neural networks for 10 epochs, and report the mean and standard deviation of $\lambda_{\min}(\text{NTK}_{(d)})$ across these runs. Table~\ref{configurations of experiments} lists the detailed configurations of the network in this experiment. All the weight matrices $\mathbf{W}$ and biases $\mathbf{b}$ are independently initialized by $\mathbf{W}^{l}\sim\mathcal{N}(0,0.01)$ and $\mathbf{b}^{l}\sim\mathcal{N}(0,0.01)$ for $l\in[L]$.

\begin{figure}[t]
	\centering
	\begin{minipage}[b]{0.48\linewidth}
		\centering
		\includegraphics[width=\linewidth]{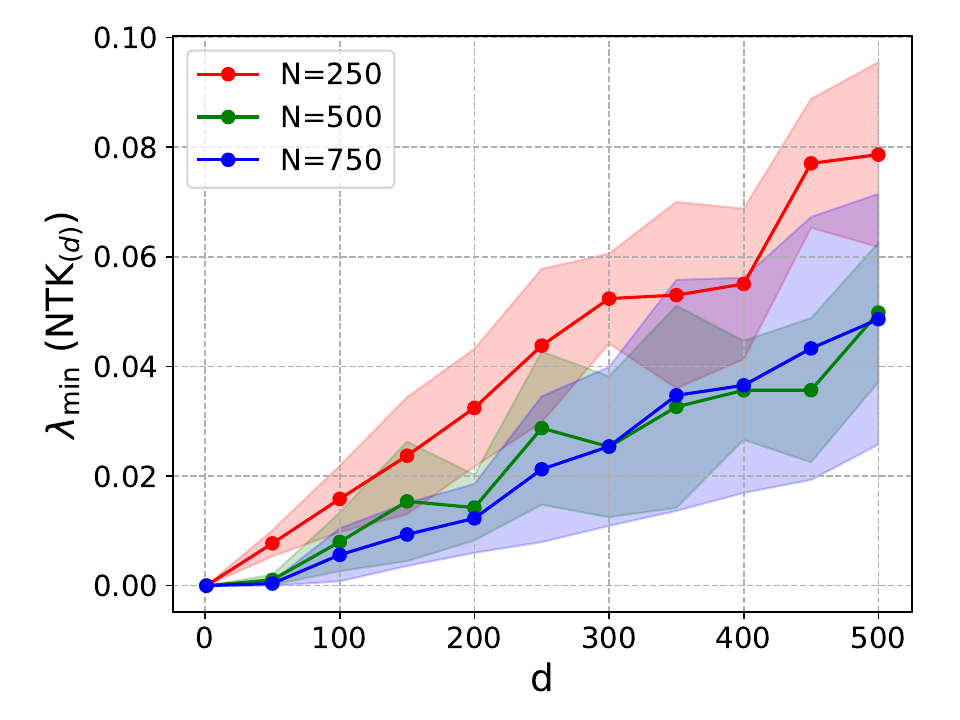}

		\small $\quad\quad$(a)
	\end{minipage}%
	\hfill
	\begin{minipage}[b]{0.48\linewidth}
		\centering
		\includegraphics[width=\linewidth]{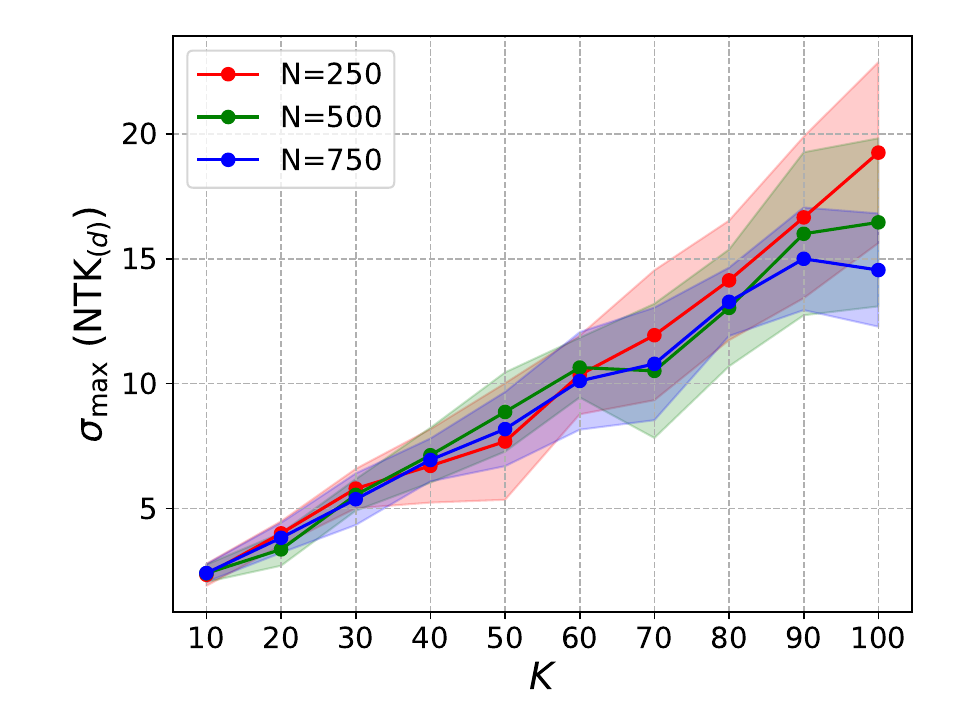}

		\small $\quad\quad\quad$(b) 
	\end{minipage}
	\caption{(a) Scaling of the smallest eigenvalue of NTK$_{(d)}$ kernel with the input dimension $d$ and (b) scaling of the largest singular value of NTK$_{(d)}$ kernel between two fixed samples $(0,1)$ and $(1,0)$ with the number of shortcut connections $K$.}
	\label{fig:bound}
\end{figure}

Figure~\ref{fig:bound}(a) displays the curves of $\lambda_{\min}(\text{NTK}_{(d)})$ with different input dimensions $d$, where the solid line indicates the mean and the shaded region represents the standard deviation. It can be observed that for all the cases of $N = 250, 500, 750$, the smallest eigenvalue of NTK$_{(d)}$ is almost linear with respect to the input dimension $d$. Thus, it validates the lower bound in Theorem~\ref{thm:smallest_eigenvalue}, which states that the smallest eigenvalue of the NTK$_{(d)}$ kernel grows at least on the order of the input dimension $d$ asymptotically. Hopefully, one can further prove that $\lambda_{\min}\left(\text{NTK}_{(d)}\right) \leq \mathcal{O}(d)$, so that the smallest eigenvalue of NTK$_{(d)}$ can be precisely depicted by the input dimension $d$. We leave this as our future work.

We also investigate the largest singular value of the NTK$_{(d)}$ kernel, with experimental configurations similar to those employed when we investigate its smallest eigenvalue. Inspired by Theorem~\ref{thm:largest_eigenvalue}, we fix the input dimension $d$ as $2$, and aim to capture the relationship between its largest singular value and the number of shortcut connections $K$. Since the upper bound in Theorem~\ref{thm:largest_eigenvalue} applies to any two input samples, we fix $(0,1)$ and $(1,0)$ as the reference points for computing the NTK$_{(d)}$ kernel.

Figure~\ref{fig:bound}(b) displays the curves of $\sigma_{\max}(\text{NTK}_{(d)})$ between two fixed samples $(0,1)$ and $(1,0)$ with different numbers of shortcut connections $K$, where the solid line indicates the mean and the shaded region represents the standard deviation. It can be observed that for all the cases of $N = 250, 500, 750$, the largest singular value of NTK$_{(d)}$ is almost linear with respect to the number of shortcut connections $K$. It is consistent with the upper bound in Theorem~\ref{thm:largest_eigenvalue}, which states that the largest singular value of the NTK$_{(d)}$ kernel grows at most quadratically with the number of shortcut connections $K$.

\begin{figure}[t]
	\centering
	\begin{minipage}[b]{0.48\linewidth}
		\centering
		\includegraphics[width=\linewidth]{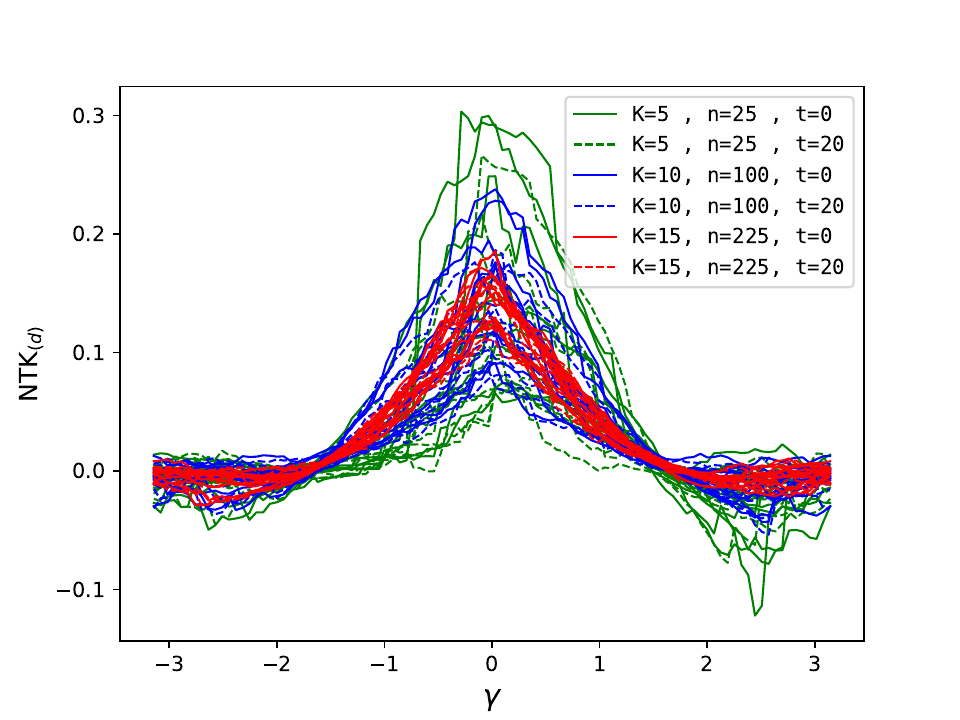}
		\small (a)
	\end{minipage}%
	\hfill
	\begin{minipage}[b]{0.48\linewidth}
		\centering
		\includegraphics[width=\linewidth]{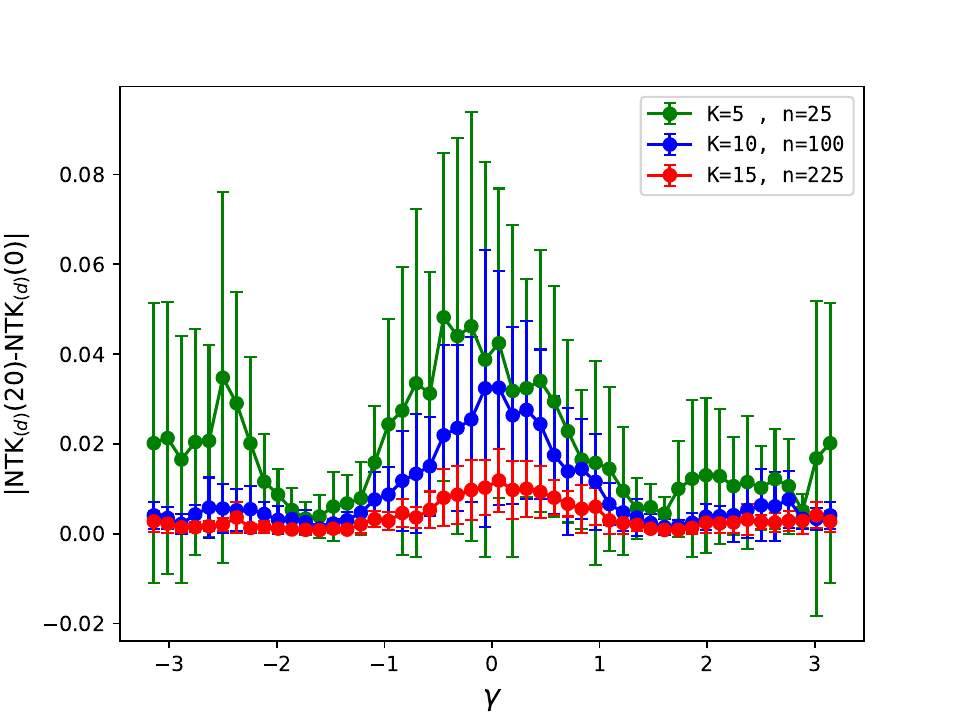}
		\small $\quad$(b) 
	\end{minipage}
	
	\caption{(a) The curves of NTK$_{(d)}$ with different $K$ at $t=0$ and $t=20$ and (b) the error bar plot of $|\text{NTK}_{(d)}(20)-\text{NTK}_{(d)}(0)|$ with different $K$ as $\gamma$ varies from $-\pi$ to $\pi$.}
	\label{fig:invariance}
\end{figure}

\subsection{Invariance during Training}\label{subsec: exp_invariance}
From Theorem~\ref{thm:constant}, the NTK$_{(d)}$ kernel remains invariant during training. We empirically validate this by following the experimental configurations of Jacot et al.~\citep{WideNTK2021}. 

The data set is simulated, with two-dimensional input on the unit circle and single-dimensional output. Specifically, for $\gamma\in[-\pi,\pi]$, the input is $(\cos\gamma,\sin\gamma)$, and the corresponding label is $\cos\gamma\sin\gamma$. We train several shortcut-related deep neural networks with different $K$ for $20$ epochs. To coincide with the condition $n=\Omega(K^{2/\lambda})$ in Theorem~\ref{thm:constant}, we force $n = K^2$. Table~\ref{configurations of experiments} lists the detailed configurations of the network in this experiment. Here, we employ the scaling factor $1/\sqrt{n_l}$ before the weight matrices following Eq.~\eqref{eq:forward}. Based on this design, the weight matrices and the biases are independently initialized with Gaussian distributions, that is $\mathbf{W}^l\sim\mathcal{N}(0,1)$ and $\boldsymbol{b}^l\sim\mathcal{N}(0,1)$ for $l\in[L]$. Each network is trained $10$ times with different initializations to account for variance in performance.

The experimental results are plotted in Figure~\ref{fig:invariance}. Figure~\ref{fig:invariance}(a) displays the curves of NTK$_{(d)}$ with different $K$ at $t=0$ and $t=20$. The curves of NTK$_{(d)}$  are plotted between a fixed point $\boldsymbol{x}_\text{fixed} = (1,0)$ and $\boldsymbol{x}=(\cos\gamma,\sin\gamma)$ for $\gamma\in[-\pi,\pi]$. We conduct $10$ trials for each $K$ to observe general patterns and trends. As the network becomes deeper, the curves of NTK$_{(d)}$ show less variance and are smoother. Further, we define an indicator $|\text{NTK}_{(d)}(20)-\text{NTK}_{(d)}(0)|$ to indicate the gap between NTK$_{(d)}(t=0)$ and NTK$_{(d)}(t=20)$. Figure~\ref{fig:invariance}(b) displays the error bar of $|\text{NTK}_{(d)}(20)-\text{NTK}_{(d)}(0)|$ with different $K$ over $\gamma \in [-\pi,\pi]$. We also conduct $10$ trials for each $K$ to compute the mean and standard deviation of the defined indicator. There are two obvious findings: (1) The value of $|\text{NTK}_{(d)}(20)-\text{NTK}_{(d)}(0)|$ is closer to zero as the network becomes deeper. We also calculate the mean across different $\gamma$ for each $K$, which indicate $0.019,0.009,0.004$ for $K={5,10,15}$, respectively. From these observations, the NTK$_{(d)}$ kernel at epoch $20$ stays closer to the NTK$_{(d)}$ kernel at initialization when the network becomes deeper, implying that NTK$_{(d)}$ maintains invariance during training. (2) The standard deviation of $|\text{NTK}_{(d)}(20)-\text{NTK}_{(d)}(0)|$ is smaller as the network becomes deeper, revealing that NTK$_{(d)}$ remains more stable and invariant during training with larger network depth. The above findings confirm the effectiveness of Theorem~\ref{thm:constant}.

\subsection{Discussions on the Experiments}\label{discussions_on_exp}
This subsection provides discussions on the experiments. The first is about the setting of parameters in experiments. All of our experiments in Section~\ref{Experiments} follow the parameter settings used in prior studies under the neural kernel regime, and we specify
them at the beginning of each subsection. However, we have to point out that for both NTK$_{(w)}$ kernel and NTK$_{(d)}$ kernel, there exists an inevitable gap between theoretical results and empirical validation in practice. The theoretical analysis often assumes infinitely wide or infinitely deep neural networks, which are infeasible in real-world experiments. For the NTK$_{(w)}$ kernel, a width around $1000$ can be regarded as sufficiently large to approximate the infinite-width regime~\citep{pmlr-v139-nguyen21g}. In our experiments, we slightly reduce this value to enable a fair comparison between the NTK$_{(w)}$ kernel and NTK$_{(d)}$ kernel. As for the NTK$_{(d)}$ kernel, the largest depth in our experiments reaches $150$, which can be regarded as sufficiently large for two reasons: (1) Several works formally show that increasing network depth yields far greater gains in complexity and expressiveness than merely increasing width~\citep{eldan2016power,montufar2014number}. Thus, a depth of $150$ is roughly comparable in representational capacity to a width exceeding $1000$, and thus adequate for our study. Moreover, a depth of $150$ is already on par with standard deep architectures such as ResNet-$152$~\citep{he2016deep}, and further increasing depth may introduce optimization or stability issues rather than performance benefits. (2) Previous studies that attempt to extend the width-induced NTK framework to account for depth typically regard depths no greater than $200$ as effectively infinite in numerical experiments~\citep{seleznova2022neural}, and we follow this setting.

The second is about possible issues we have avoided. It is known that for deep neural networks, there exists only a narrow range of weight initialization that can avoid vanishing gradients~\citep{schoenholz2016deep}. Therefore, proper initialization must be carefully considered in this paper. As stated in Definition~\ref{def:well_posed}, satisfying the stable-pertinent property for the weights is fully compatible with Gaussian initialization, such as the commonly used Xavier initialization~\citep{glorot2010understanding} and He initialization~\citep{he2015delving}. Therefore, in our experiments, we adopt ReLU activation with He initialization and train the networks using vanilla gradient descent, which are standard practices in deep learning. Under this setting, gradient vanishing or explosion can be effectively controlled~\citep{he2016deep}. Moreover, Theorems~\ref{thm:smallest_eigenvalue} and~\ref{thm:largest_eigenvalue} guarantee that the NTK$_{(d)}$ kernel itself neither degenerates to zero nor explode. In Experiment~\ref{subsec: lower_bound_exp}, we have further validated this point empirically.

\section{Conclusions and Prospects}\label{Conclusions}
In this paper, we proposed a depth-induced NTK kernel based on a specific shortcut-related network architecture. Provided the depth-induced neural kernel, we further conclude three characterizations: (1) the lower bound on the smallest eigenvalue grows at least on the order of the input dimension; (2) the upper bound on the largest singular value of the NTK$_{(d)}$ kernel grows at most quadratically with the number of shortcut connections; and (3) the NTK$_{(d)}$ kernel remains invariant under mild conditions. Furthermore, we conduct kernel regression experiments on both simulated and real-world data sets, showing that the proposed NTK$_{(d)}$ kernel has comparable capability to the traditional width-induced NTK$_{(w)}$ kernel. We empirically investigate the effects of the separation constant $\hbar$ and the number of shortcut connections $K$ on the behaviors of NTK$_{(d)}$, in which $K$ has a greater effect than $\hbar$, coinciding with our theory. Also, we validate the established bounds on the eigenvalue or singular value of NTK$_{(d)}$ and confirm that NTK$_{(d)}$ remains invariant during training in practice. Our findings greatly extend the scope of the neural kernel theory and provide in-depth insights of over-parameterized models. 

The significance of the NTK$_{(d)}$ kernel can be interpreted at three levels. (1) The NTK$_{(d)}$ kernel effectively breaks through the limitation of infinite width, and addresses the severe performance degradation suffered by the traditional NTK$_{(w)}$ kernel as the network becomes deeper~\citep{huang2020dynamics}. Thereby, our work greatly extends the applicability of the NTK framework to account for the analysis of deep neural networks. (2) Eigenvalue analysis of NTK$_{(d)}$ enables the further study of the generalization properties of deep learning from the perspective of kernel methods~\citep{arora2019fine,chen2020generalized,simon2019gradient}, a task that was considered difficult and important to achieve~\citep{zhang2017understanding, zhou2021over}. (3) By proposing the NTK$_{(d)}$ kernel, a tool inherently suited for analyzing deep neural networks, we aim to facilitate an in-depth understanding of deep learning.

This research suggests several important avenues for future work. One attractive future study is to enhance the applicability of the NTK$_{(d)}$ kernel. As stated in Theorem~\ref{thm:NTK_depth}, we have provided a theoretical guarantee and feasibility justification for the depth-induced NTK$_{(d)}$ kernel. Although our current analysis is conducted based on a specific shortcut network architecture, the proposed methodology is expected to generalize. Notably, the only architectural restriction is the presence of weak dependence. Thereby, it is interesting to extend this investigation to a broader class of architectures in future work. Additionally, we have theoretically established in Theorem~\ref{thm:smallest_eigenvalue} that the smallest eigenvalue of NTK$_{(d)}$ admits a linear lower bound with respect to the input dimension $d$. Interestingly, as shown in Subsection~\ref{subsec: lower_bound_exp}, the empirical results further indicate that the eigenvalue itself exhibits an overall linear dependence on $d$. This suggests that the upper bound might also scale linearly with $d$, i.e., $\lambda_{\min}(\text{NTK}_{(d)}) \leq \mathcal{O}(d)$, which we leave as future work. Finally, our regression experiments demonstrate that the depth-induced NTK$_{(d)}$ kernel achieves strong performance with deep neural networks, effectively addressing the severe performance degradation suffered by the traditional width-induced NTK$_{(w)}$ kernel as the network becomes deeper~\citep{huang2020dynamics}. Future research could further investigate the underlying mystery of why our depth-induced NTK$_{(d)}$ kernel overcomes this degradation.

\section*{Acknowledgments and Disclosure of Funding}
The research was funded by multiple grants. Yong-Ming Tian was supported by the National Natural Science Foundation of China Young Students Basic Research Project (625B1014). Shao-Qun Zhang was supported by the Natural Science Foundation of China (62406138).

\bibliography{JMref}
\bibliographystyle{plain}

\newpage
\appendix
{\Large\bfseries Appendix} \\~\\
This appendix provides the supplementary materials for our work ``Depth-induced NTK: Bridging Over-parameterized Neural Networks and Deep Neural Kernels".

\section{Proof of Lemma~\ref{lemma:main_prod_sum_weak}}\label{appendix:proof_of_prod_sum_weak}
{\textit{Proof.}}
Without loss of generality, we prove Lemma~\ref{lemma:main_prod_sum_weak} by the first pair of conditions of Lemma~\ref {lemma:main_weak} when stating a sequence is weakly dependent. We first prove that $\{A^i\}_{i=1}^{t}$ is weakly dependent. As the sequences $\{X^i\}_{i=1}^{t}$ and $\{Y^i\}_{i=1}^{t}$ are weakly dependent sequences constructed from the sequences $\{\tilde{X}^i\}_{i=1}^{t^\prime}$ and $\{\tilde{Y}^i\}_{i=1}^{t^\prime}$, by Lemma~\ref{lemma:main_weak}, there exist constants $C_1$, $C_2$ and statistical variables $R_1$, $R_2$ such that for any $l\in\mathbb{N}, \hbar\in \mathbb{N}^+$ satisfying $1 \leq l+\hbar \leq t$, the following conditions hold
\[
\left\{
\begin{aligned}
	~&\left\|{\partial \tilde{X}^{l+\hbar}}/{\partial \tilde{X}^l}\right\|_s \leq C_1^\hbar\quad \text{and}\quad {\left|\tilde{X}_p^{l+\hbar}\tilde{X}_q^l\right|}/{\left|\tilde{X}_q^{l}\right|} \leq R_1~ C_1^\hbar\ , \\
	&\left\|{\partial \tilde{Y}^{l+\hbar}}/{\partial \tilde{Y}^l}\right\|_s \leq C_2^\hbar\quad \text{and}\quad {\left|\tilde{Y}_p^{l+\hbar}\tilde{Y}_q^l\right|}/{\left|\tilde{Y}_q^{l}\right|} \leq R_2~ C_2^\hbar\ .
\end{aligned}
\right.
\]
where $\tilde{X}_p^{l+\hbar}$ and $\tilde{X}_q^{l}$ separately denote any element in $\tilde{X}^{l+\hbar}$ and $\tilde{X}^{l}$, and $\tilde{Y}_p^{l+\hbar}$ and $\tilde{Y}_q^{l}$ separately denote any element in $\tilde{Y}^{l+\hbar}$ and $\tilde{Y}^{l}$.

The variables in $\{\tilde{X}^i\}_{i=1}^{t}$ and $\{\tilde{Y}^i\}_{i=1}^{t}$ must have finite spectral norm, upper bounded by a universal Constant $C$. Without loss of generality, we suppose $C, C_1, C_2 \geq 1$. By the independence between $\tilde{X}^i$ and $\tilde{Y}^i$ for any $i\in[t]$, we have

\[
\left\{
\begin{aligned}
	~&\left\|{\partial \tilde{A}^{l+\hbar}}/{\partial \tilde{X}^l}\right\|_s \leq \|\tilde{Y}^{l+\hbar}\|_s~\left\|{\partial \tilde{X}^{l+\hbar}}/{\partial \tilde{X}^l}\right\|_s \leq C C_1^\hbar \leq \left(C C_1 C_2\right)^\hbar\ , \\
	&\left\|{\partial \tilde{A}^{l+\hbar}}/{\partial \tilde{Y}^l}\right\|_s \leq \|\tilde{X}^{l+\hbar}\|_s~\left\|{\partial \tilde{Y}^{l+\hbar}}/{\partial \tilde{Y}^l}\right\|_s \leq C C_2^\hbar \leq \left(C C_1 C_2\right)^\hbar\ .
\end{aligned}
\right.
\]
Moreover, the following holds
\[
{\left|\tilde{A}_p^{l+\hbar}\tilde{A}_q^l\right|}/{\left|\tilde{A}_q^l\right|}
= {\left|\tilde{X}_p^{l+\hbar}\tilde{Y}_p^{l+\hbar}\tilde{X}_q^l\tilde{Y}_q^l\right|}/{\left|\tilde{X}_q^l\tilde{Y}_q^l\right|}
\leq R_1 C_1^\hbar R_2 C_2^\hbar \leq R_1 R_2 \left(C C_1 C_2\right)^\hbar\ .
\]
Thus, the sequence $\{A^i\}_{i=1}^{t}$ is a weakly dependent sequence constructed from $\{\tilde{A}^i\}_{i=1}^{t^\prime}$.

Similarly, we prove that the sequence $\{B^i\}_{i=1}^{t}$ is a weakly dependent sequence constructed from $\{\tilde{B}^i\}_{i=1}^{t^\prime}$. We verify the conditions in Lemma~\ref{lemma:main_prod_sum_weak}. Without loss of generality, we suppose $C_1, C_2 \geq 1$. Firstly, by the independence of $\tilde{X}^i$ and $\tilde{Y}^i$ for any $i \in [t^\prime]$, we have
\[
\left\{~
\begin{aligned}
	&\left\|{\partial \tilde{B}^{l+\hbar}}/{\partial \tilde{X}^l}\right\|_s = \left\|{\partial \tilde{X}^{l+\hbar}}/{\partial \tilde{X}^l}\right\|_s \leq C_1^\hbar \leq \left(C_1 + C_2\right)^\hbar\ , \\
	&\left\|{\partial \tilde{B}^{l+\hbar}}/{\partial \tilde{Y}^l}\right\|_s = \left\|{\partial \tilde{Y}^{l+\hbar}}/{\partial \tilde{Y}^l}\right\|_s \leq C_2^\hbar \leq \left(C_1 + C_2\right)^\hbar\ .
\end{aligned}
\right.
\]
Then, one has
\[
\frac{\left|\tilde{B}_p^{l+\hbar}\right|}{\left|\tilde{B}_q^l\right|}
= \frac{\left|\tilde{X}_p^{l+\hbar}+\tilde{Y}_p^{l+\hbar}\right|}{\left|\tilde{X}_q^l+\tilde{Y}_q^l\right|}
\leq \frac{\left|\tilde{X}_p^{l+\hbar}\right|+\left|\tilde{Y}_p^{l+\hbar}\right|}{\left|\tilde{X}_q^l+\tilde{Y}_q^l\right|} \leq \frac{R_1 C_1^\hbar + R_2 C_2^\hbar}{\left|\tilde{X}_q^l+\tilde{Y}_q^l\right|}\ .
\]
Moreover, we have
\[
{\left|\tilde{B}_p^{l+\hbar}\tilde{B}_q^l\right|}/{\left|\tilde{B}_q^l\right|} \leq \max\{R_1,R_2\} ~\left(C_1 + C_2\right)^\hbar\ .
\]
Thus, the conditions in Lemma~\ref{lemma:main_weak} are satisfied, concluding the proof. $\hfill\blacksquare$

\section{Full Proof of Theorem~\ref{thm:NTK_depth}}\label{fullproof:3.1}
{\bf Statement of Theorem~\ref{thm:NTK_depth}.} Provided the shortcut-related network defined by Eqs.~\eqref{eq:forward} and~\eqref{eq:shortcut}, if the following conditions hold
\begin{itemize}
	\vspace{-12pt}
	\item[(1)] $\phi$ is well-posed,
	\vspace{-5pt}
	\item[(2)] the weight matrices of each layer are stable-pertinent for $\phi$, that is, $\mathbf{W}^l \in SP(\phi)$ for $\forall~ l\in [L]$,
	\vspace{-8pt}
\end{itemize}
there derives an NTK kernel that converges to a Gaussian distribution in the limit of the number of shortcut connections as well as the network depth going to infinity, that is, $K\rightarrow +\infty$ as well as $L\rightarrow +\infty$. Formally, for any inputs $\boldsymbol{x}$ and $\boldsymbol{x}^\prime$, we have
\begin{equation}\label{sum_of_NTK_kappa_hbar}
	\text{NTK}_{(d)} (\boldsymbol{x}, \boldsymbol{x}') = \sum_{\kappa^\prime = 0}^{K}\text{NTK}_{(d)}^{\kappa^\prime\hbar} (\boldsymbol{x}, \boldsymbol{x}') \ ,
\end{equation}
with
\begin{equation} \label{definition_form} 
	\text{NTK}_{(d)}^{\kappa^\prime\hbar} (\boldsymbol{x}, \boldsymbol{x}')\triangleq \left\langle\text{D}_b^{n_{\kappa^\prime\hbar-1}} \{{\mathbf{W}^{\kappa^\prime\hbar}}^\top \}\frac{\partial f(\boldsymbol{x})}{\partial \mathbf{W}^{\kappa^\prime\hbar}}, \text{D}_b^{n_{\kappa^\prime\hbar-1}}\{{\mathbf{W}^{\kappa^\prime\hbar}}^\top\}\frac{\partial f(\boldsymbol{x}^\prime)}{\partial \mathbf{W}^{\kappa^\prime\hbar}}\right\rangle 
\end{equation}
and
\begin{equation} \label{expanded_form} 
	\text{NTK}_{(d)}^{\kappa^\prime\hbar} (\boldsymbol{x}, \boldsymbol{x}') = \frac{1}{{M_{\boldsymbol{z}}}}\left\langle\boldsymbol{z}^{\kappa^\prime\hbar-1}(\boldsymbol{x}),\boldsymbol{z}^{\kappa^\prime\hbar-1}(\boldsymbol{x}^\prime)\right\rangle \left\langle\Delta^{\kappa^\prime\hbar}(\boldsymbol{x}),\Delta^{\kappa^\prime\hbar}(\boldsymbol{x}^\prime)\right\rangle  \ ,
\end{equation}
where $\Delta^{\kappa^\prime\hbar}(\boldsymbol{x}) = \sum_{\kappa = \kappa^\prime}^{K}\left[\prod_{i=\kappa^\prime\hbar}^{\kappa\hbar}\left(\hat{\mathbf{W}}^{i}\right)^\top\mathbf{D}^{i}(\boldsymbol{x})\right]~{\mathbf{J}^{\kappa\hbar}}^\top$ and~ $\mathbf{D}^i(\boldsymbol{x}) = \text{diag}\{\dot{\phi}(\hat{\mathbf{W}}^i\boldsymbol{z}^{i-1}(\boldsymbol{x}))\}$\ .

Before starting with the proof of Theorem~\ref{thm:NTK_depth}, we make some important preparations in the following subsections. Subsection~\ref{subsec:expanded_form_of_delta} proves that Eq.~\eqref{definition_form} can be expanded into Eq.~\eqref{expanded_form}. Subsection~\ref{A useful lemma about the largest singular value of a matrix} gives a useful lemma about the largest singular value of a matrix.

\subsection{Proof of the Expanded Form of NTK\texorpdfstring{$_{(d)}^{\kappa^\prime\hbar}$}{d kappa prime hbar}}\label{subsec:expanded_form_of_delta}
{\textit{Proof.}} We firstly calculate the gradient of the weight parameters in the $\kappa^\prime\hbar$-th layer $\partial f(\boldsymbol{x})/\partial \mathbf{W}^{\kappa^\prime\hbar}$ for $\kappa^\prime \in[K]\ .$ Notice that $\mathbf{W}^{\kappa^\prime\hbar}$ are independently initialized, so if we calculate $\partial \boldsymbol{z}^{\kappa\hbar}/\partial \mathbf{W}^{\kappa^\prime\hbar}$ where $\boldsymbol{z}^{\kappa\hbar}$ is before $\mathbf{W}^{\kappa^\prime\hbar}$, namely $\kappa < \kappa^\prime$, the term is zero. When there is no ambiguity, we simply use $\boldsymbol{z}^{\kappa\hbar}$ to denote $\boldsymbol{z}^{\kappa\hbar}(\boldsymbol{x})$.
The output of the concerned network is computed by
\begin{equation}\label{appendix:output}
	f(\boldsymbol{x}) = \frac{1}{\sqrt{M_{\boldsymbol{z}}}} \sum_{\kappa=0}^{K} \mathbf{J}^{\kappa\hbar} \boldsymbol{z}^{\kappa\hbar}(\boldsymbol{x}) \ .
\end{equation}
Let $\mathbf{D}^i$ denote $\text{diag}\{\dot{\phi}(\hat{\mathbf{W}}^i\boldsymbol{z}^{i-1})\}$. By substituting Eq.~\eqref{appendix:output} into the parameter gradient, we have
\begin{equation}\label{partial_f}
	\frac{\partial f(\boldsymbol{x})}{\partial \mathbf{W}^{\kappa^\prime\hbar}} 
	= \frac{1}{\sqrt{M_{\boldsymbol{z}}}} \sum\nolimits_{\kappa = \kappa^\prime}^{K}\frac{  \partial \boldsymbol{z}^{\kappa\hbar}}{\partial \mathbf{W}^{\kappa^\prime\hbar}}~(\mathbf{I}_1\otimes{\mathbf{J}^{\kappa\hbar}}^\top)
	= \frac{1}{\sqrt{M_{\boldsymbol{z}}}} \sum\nolimits_{\kappa = \kappa^\prime}^{K}\frac{  \partial \boldsymbol{z}^{\kappa\hbar}}{\partial \mathbf{W}^{\kappa^\prime\hbar}}~{\mathbf{J}^{\kappa\hbar}}^\top\ ,
\end{equation}
where
\begin{align}\label{partial_z}
	\frac{  \partial \boldsymbol{z}^{\kappa\hbar}}{\partial \mathbf{W}^{\kappa^\prime\hbar}}\nonumber
	&=\frac{\partial\phi(\hat{\mathbf{W}}^{\kappa\hbar}\phi(\hat{\mathbf{W}}^{\kappa\hbar-1}\cdots\phi(\hat{\mathbf{W}}^{\kappa^\prime\hbar}\boldsymbol{z}^{\kappa^\prime\hbar-1})))}{\partial \mathbf{W}^{\kappa^\prime\hbar}}\nonumber \\
	&=  \frac{\partial\hat{\mathbf{W}}^{\kappa\hbar}\phi(\hat{\mathbf{W}}^{\kappa\hbar-1}\cdots\phi(\hat{\mathbf{W}}^{\kappa^\prime\hbar}\boldsymbol{z}^{\kappa^\prime\hbar-1}))}{\partial \mathbf{W}^{\kappa^\prime\hbar}} \frac{\partial\phi(\hat{\mathbf{W}}^{\kappa\hbar}\phi(\hat{\mathbf{W}}^{\kappa\hbar-1}\cdots\phi(\hat{\mathbf{W}}^{\kappa^\prime\hbar}\boldsymbol{z}^{\kappa^\prime\hbar-1})))}{\partial\hat{\mathbf{W}}^{\kappa\hbar}\phi(\hat{\mathbf{W}}^{\kappa\hbar-1}\cdots\phi(\hat{\mathbf{W}}^{\kappa^\prime\hbar}\boldsymbol{z}^{\kappa^\prime\hbar-1}))}\nonumber \\
	&= \frac{\partial\phi(\hat{\mathbf{W}}^{\kappa\hbar-1}\cdots\phi(\hat{\mathbf{W}}^{\kappa^\prime\hbar}\boldsymbol{z}^{\kappa^\prime\hbar-1}))}{\partial \mathbf{W}^{\kappa^\prime\hbar}}\frac{\partial\hat{\mathbf{W}}^{\kappa\hbar}\phi(\hat{\mathbf{W}}^{\kappa\hbar-1}\cdots\phi(\hat{\mathbf{W}}^{\kappa^\prime\hbar}\boldsymbol{z}^{\kappa^\prime\hbar-1}))}{\partial\phi(\hat{\mathbf{W}}^{\kappa\hbar-1}\cdots\phi(\hat{\mathbf{W}}^{\kappa^\prime\hbar}\boldsymbol{z}^{\kappa^\prime\hbar-1}))} \mathbf{D}^{\kappa\hbar} \nonumber \\
	&= \frac{\partial\phi(\hat{\mathbf{W}}^{\kappa\hbar-1}\cdots\phi(\hat{\mathbf{W}}^{\kappa^\prime\hbar}\boldsymbol{z}^{\kappa^\prime\hbar-1}))}{\partial \mathbf{W}^{\kappa^\prime\hbar}} \left(\mathbf{I}_1\otimes\left(\hat{\mathbf{W}}^{\kappa\hbar}\right)^\top\right) \mathbf{D}^{\kappa\hbar} \nonumber \\
	&= \frac{\partial\phi(\hat{\mathbf{W}}^{\kappa\hbar-1}\cdots\phi(\hat{\mathbf{W}}^{\kappa^\prime\hbar}\boldsymbol{z}^{\kappa^\prime\hbar-1}))}{\partial \mathbf{W}^{\kappa^\prime\hbar}} \left(\hat{\mathbf{W}}^{\kappa\hbar}\right)^\top \mathbf{D}^{\kappa\hbar} \nonumber \\
	&= \frac{\partial \hat{\mathbf{W}}^{\kappa^\prime\hbar}\boldsymbol{z}^{\kappa^\prime\hbar-1}}{\mathbf{W}^{\kappa^\prime\hbar}} \frac{\partial\phi( \hat{\mathbf{W}}^{\kappa^\prime\hbar}\boldsymbol{z}^{\kappa^\prime\hbar-1})}{\partial \mathbf{W}^{\kappa^\prime\hbar}\boldsymbol{z}^{\kappa^\prime\hbar-1}} \prod\nolimits_{i=\kappa^\prime\hbar+1}^{\kappa\hbar}\left(\hat{\mathbf{W}}^i\right)^\top\mathbf{D}^i\nonumber \\
	&= \frac{1}{\sqrt{n_{\kappa^\prime\hbar}}}(\boldsymbol{z}^{\kappa^\prime\hbar-1}\otimes\mathbf{I}_{n_{\kappa^\prime\hbar}}) \mathbf{D}^{\kappa^\prime\hbar} \prod\nolimits_{i=\kappa^\prime\hbar+1}^{\kappa\hbar}\left(\hat{\mathbf{W}}^i\right)^\top\mathbf{D}^i \nonumber \\
	&= \frac{1}{\sqrt{n_{\kappa^\prime\hbar}}}(\boldsymbol{z}^{\kappa^\prime\hbar-1}\otimes\mathbf{I}_{n_{\kappa^\prime\hbar}}) \left[\mathbf{I}_1\otimes\left( \mathbf{D}^{\kappa^\prime\hbar} \prod\nolimits_{i=\kappa^\prime\hbar+1}^{\kappa\hbar}\left(\hat{\mathbf{W}}^i\right)^\top \mathbf{D}^i \right)\right]\nonumber \\
	&= \frac{1}{\sqrt{n_{\kappa^\prime\hbar}}}(\boldsymbol{z}^{\kappa^\prime\hbar-1}~\mathbf{I}_1)\otimes\left[\mathbf{I}_{n_{\kappa^\prime\hbar}}\left(\mathbf{D}^{\kappa^\prime\hbar} \prod\nolimits_{i=\kappa^\prime\hbar+1}^{\kappa\hbar}\left(\hat{\mathbf{W}}^i\right)^\top \mathbf{D}^i \right)\right]\nonumber \\
	&= \frac{\boldsymbol{z}^{\kappa^\prime\hbar-1}}{\sqrt{n_{\kappa^\prime\hbar}}}\otimes\left[\text{diag}\{\dot{\phi}(\hat{\mathbf{W}}^{\kappa^\prime\hbar}\boldsymbol{z}^{\kappa^\prime\hbar-1})\}\prod\nolimits_{i=\kappa^\prime\hbar+1}^{\kappa\hbar}\left(\hat{\mathbf{W}}^i\right)^\top \mathbf{D}^i\right]  \ , 
\end{align}
By inserting Eq.~\eqref{partial_z} into Eq.~\eqref{partial_f}, and by expanding $\boldsymbol{z}^{\kappa^\prime\hbar-1}$ into component form, we have
\begin{align}\label{eq:derivative}
	\frac{\partial f(\boldsymbol{x})}{\partial \mathbf{W}^{\kappa^\prime\hbar}} \nonumber
	&=\frac{1}{\sqrt{M_{\boldsymbol{z}}~ n_{\kappa^\prime\hbar}}}\sum_{\kappa = \kappa^\prime}^{K} \text{vec}\left(\boldsymbol{z}_j^{\kappa^\prime\hbar-1}\right)_{j=1}^{n_{\kappa^\prime\hbar-1}}\otimes\left[\mathbf{D}^{\kappa^\prime\hbar}\prod_{i=\kappa^\prime\hbar+1}^{\kappa\hbar}\left(\hat{\mathbf{W}^i}\right)^\top\mathbf{D}^i~{\mathbf{J}^{\kappa\hbar}}^\top\right]\nonumber \\
	&=\frac{1}{\sqrt{M_{\boldsymbol{z}}~ n_{\kappa^\prime\hbar}}}\sum_{\kappa = \kappa^\prime}^{K} \text{vec}\left(\boldsymbol{z}_j^{\kappa^\prime\hbar-1}\left[\mathbf{D}^{\kappa^\prime\hbar}\prod_{i=\kappa^\prime\hbar+1}^{\kappa\hbar}\left(\hat{\mathbf{W}^i}\right)^\top\mathbf{D}^i~ {\mathbf{J}^{\kappa\hbar}}^\top\right]\right)_{j=1}^{n_{\kappa^\prime\hbar-1}}\nonumber\ .\\
\end{align}
Let $\mathbf{F}^{\kappa\hbar}$ denote the whole part
\[
\left[\mathbf{D}^{\kappa^\prime\hbar}\prod_{i=\kappa^\prime\hbar+1}^{\kappa\hbar}\left(\hat{\mathbf{W}}^i\right)^\top\mathbf{D}^i~ {\mathbf{J}^{\kappa\hbar}}^\top\right]
\]
for simplicity. Thus, the gradient can be rewritten as
\begin{equation}\label{eq:simplified_gradient}
	\frac{\partial f(\boldsymbol{x})}{\partial \mathbf{W}^{\kappa^\prime\hbar}} = \frac{1}{\sqrt{M_{\boldsymbol{z}}~n_{\kappa^\prime\hbar}}}\sum_{\kappa = \kappa^\prime}^{K}\text{vec}\left(\boldsymbol{z}_j^{\kappa^\prime\hbar-1}\mathbf{F}^{\kappa\hbar}(\boldsymbol{x})\right)_{j=1}^{n_{\kappa^\prime\hbar-1}}\ .
\end{equation}
Then, by substituting Eq.~\eqref{eq:simplified_gradient} into Eq.~\eqref{definition_form}, we can expand NTK$_{(d)}^{\kappa^\prime\hbar}$ as
\[
\resizebox{0.99\textwidth}{!}{
	$\begin{aligned}
		\text{NTK}_{(d)}^{\kappa^\prime\hbar}
		&=\left\langle\text{D}_b^{n_{\kappa^\prime\hbar-1}} \{{\mathbf{W}^{\kappa^\prime\hbar}}^\top \}\frac{\partial f(\boldsymbol{x})}{\partial \mathbf{W}^{\kappa^\prime\hbar}},\text{D}_b^{n_{\kappa^\prime\hbar-1}} \{{\mathbf{W}^{\kappa^\prime\hbar}}^\top \}\frac{\partial f(\boldsymbol{x}^\prime)}{\partial \mathbf{W}^{\kappa^\prime\hbar}}\right\rangle\nonumber \\
		&= \left\langle \frac{\text{D}_b^{n_{\kappa^\prime\hbar-1}} \{{\mathbf{W}^{\kappa^\prime\hbar}}^\top \}}{\sqrt{M_{\boldsymbol{z}}~ n_{\kappa^\prime\hbar}}}\sum_{\kappa = \kappa^\prime}^{K}\text{vec}\left(\boldsymbol{z}_j^{\kappa^\prime\hbar-1}(\boldsymbol{x})\mathbf{F}^{\kappa\hbar}(\boldsymbol{x})\right)_{j=1}^{n_{\kappa^\prime\hbar-1}}, \frac{\text{D}_b^{n_{\kappa^\prime\hbar-1}} \{{\mathbf{W}^{\kappa^\prime\hbar}}^\top \}}{\sqrt{M_{\boldsymbol{z}}~n_{\kappa^\prime\hbar}}}\sum_{\kappa = \kappa^\prime}^{K}\text{vec}\left(\boldsymbol{z}_j^{\kappa^\prime\hbar-1}(\boldsymbol{x}^\prime)\mathbf{F}^{\kappa\hbar}(\boldsymbol{x}^\prime)\right)_{j=1}^{n_{\kappa^\prime\hbar-1}}\right\rangle\nonumber \\
		&= \frac{1}{{M_{\boldsymbol{z}}}}\left\langle \text{vec}\left(\left(\hat{\mathbf{W}}^{\kappa^\prime\hbar}\right)^\top\sum_{\kappa = \kappa^\prime}^{K}\boldsymbol{z}_j^{\kappa^\prime\hbar-1}(\boldsymbol{x})\mathbf{F}^{\kappa\hbar}(\boldsymbol{x})\right)_{j=1}^{n_{\kappa^\prime\hbar-1}}, \text{vec}\left(\left(\hat{\mathbf{W}}^{\kappa^\prime\hbar}\right)^\top\sum_{\kappa = \kappa^\prime}^{K}\boldsymbol{z}_j^{\kappa^\prime\hbar-1}(\boldsymbol{x}^\prime)\mathbf{F}^{\kappa\hbar}(\boldsymbol{x}^\prime)\right)_{j=1}^{n_{\kappa^\prime\hbar-1}}\right\rangle\nonumber \\
		&= \frac{1}{{M_{\boldsymbol{z}}}}\sum_{j=1}^{n_{\kappa^\prime\hbar-1}}\left\langle\left(\hat{\mathbf{W}}^{\kappa^\prime\hbar}\right)^\top\sum_{\kappa = \kappa^\prime}^{K}\boldsymbol{z}_j^{\kappa^\prime\hbar-1}(\boldsymbol{x})\mathbf{F}^{\kappa\hbar}(\boldsymbol{x}),\left(\hat{\mathbf{W}}^{\kappa^\prime\hbar}\right)^\top\sum_{\kappa = \kappa^\prime}^{K}\boldsymbol{z}_j^{\kappa^\prime\hbar-1}(\boldsymbol{x}^\prime)\mathbf{F}^{\kappa\hbar}(\boldsymbol{x}^\prime)\right\rangle\nonumber \\
		&= \frac{1}{{M_{\boldsymbol{z}}}}\sum_{j=1}^{n_{\kappa^\prime\hbar-1}}\boldsymbol{z}_j^{\kappa^\prime\hbar-1}(\boldsymbol{x})\boldsymbol{z}_j^{\kappa^\prime\hbar-1}(\boldsymbol{x}^\prime)\left\langle\left(\hat{\mathbf{W}}^{\kappa^\prime\hbar}\right)^\top\sum_{\kappa = \kappa^\prime}^{K}\mathbf{F}^{\kappa\hbar}(\boldsymbol{x}),\left(\hat{\mathbf{W}}^{\kappa^\prime\hbar}\right)^\top\sum_{\kappa = \kappa^\prime}^{K}\mathbf{F}^{\kappa\hbar}(\boldsymbol{x}^\prime)\right\rangle\nonumber \\
		&= \frac{1}{{M_{\boldsymbol{z}}}}\left\langle\boldsymbol{z}^{\kappa^\prime\hbar-1}(\boldsymbol{x}),\boldsymbol{z}^{\kappa^\prime\hbar-1}(\boldsymbol{x}^\prime) \right\rangle\left\langle\left(\hat{\mathbf{W}}^{\kappa^\prime\hbar}\right)^\top\sum_{\kappa = \kappa^\prime}^{K}\mathbf{F}^{\kappa\hbar}(\boldsymbol{x}),\left(\hat{\mathbf{W}}^{\kappa^\prime\hbar}\right)^\top\sum_{\kappa = \kappa^\prime}^{K}\mathbf{F}^{\kappa\hbar}(\boldsymbol{x}^\prime)\right\rangle\ .
	\end{aligned} $}
\]
Let $\Delta^{\kappa^\prime\hbar}(\boldsymbol{x})$ denote $\left(\hat{\mathbf{W}}^{\kappa^\prime\hbar}\right)^\top\sum_{\kappa = \kappa^\prime}^{K}\mathbf{F}^{\kappa\hbar}(\boldsymbol{x})$, we have
\begin{equation*}
	\text{NTK}_{(d)}^{\kappa^\prime\hbar}(\boldsymbol{x},\boldsymbol{x}^\prime) = \frac{1}{{M_{\boldsymbol{z}}}}\left\langle\boldsymbol{z}^{\kappa^\prime\hbar-1}(\boldsymbol{x}),\boldsymbol{z}^{\kappa^\prime\hbar-1}(\boldsymbol{x}^\prime)\right\rangle\left\langle\Delta^{\kappa^\prime\hbar}(\boldsymbol{x}),\Delta^{\kappa^\prime\hbar}(\boldsymbol{x}^\prime)\right\rangle\ ,
\end{equation*}
which is the expanded form of NTK$_{(d)}^{\kappa^\prime\hbar}$ in Eq.~\eqref{expanded_form}. This completes the proof. \hfill$\blacksquare$

\subsection{A Useful Lemma about the Largest Singular Value of a Matrix}
\label{A useful lemma about the largest singular value of a matrix}
\begin{lemma}\label{appendix:lemma:scale_inequalities}
	Given a matrix $\mathbf{M} = [M_{ij}]\in\mathbb{R}^{n\times m}$, the inequalities below hold
	\[
	\frac{\sigma_{\max}(\mathbf{M})}{\sqrt{mn}} \leq |M_{ij}|_{\max} \leq \sqrt{\min\{m,n\}}~\sigma_{\max}(\mathbf{M})\ ,
	\]
	where $|M_{ij}|_{\max}$ denotes the largest absolute value of the elements of~$\mathbf{M}$.
\end{lemma}
{\textit{Proof.}} It is known that the Frobenius norm of a matrix $\mathbf{M}$ equals the square root of the sum of the squares of its singular values, that is,
\begin{equation}\label{element_and_singular_value}
	\|\mathbf{M}\|_{F} \triangleq {\left(\sum_{i,j}M_{ij}^2\right)}^{1/2} = {\left(\sum_k\sigma_k^2\right)}^{1/2} 
\end{equation}
By separately scaling the expressions in Eq.~\eqref{element_and_singular_value}, we have
\[
\left\{
\begin{aligned}
	|M_{ij}|_{\max} &\leq {\left(\sum_{i,j}M_{ij}^2\right)}^{1/2} \leq \sqrt{mn}~|M_{ij}|_{\max}\ , \\
	\sigma_{\max}(\mathbf{M}) &\leq {\left(\sum_k\sigma_k^2\right)}^{1/2} \leq \sqrt{\min\{m,n\}}~\sigma_{\max}(\mathbf{M})\ ,
\end{aligned}
\right.
\]
which concludes the proof. $\hfill\blacksquare$

\subsection{Finishing the Proof of Theorem~\ref{thm:NTK_depth}}\label{finishing_the_proof_of_theorem_3.1}
{\textit{Proof.}} Our goal is to prove the weak dependence of the sequence $\{\text{NTK}_{(d)}^{\kappa^\prime\hbar}\}_{\kappa^\prime=1}^{K}$. By the expanded form of NTK$_{(d)}^{\kappa^\prime\hbar}$ in Eq.~\eqref{expanded_form}
and Lemma~\ref{lemma:main_prod_sum_weak}, it suffices to prove that the sequences 
\[\{\boldsymbol{z}^{\kappa^\prime\hbar-1}(\boldsymbol{x})\}_{\kappa^\prime=1}^{K}\quad \text{and}\quad \{\Delta^{\kappa^\prime\hbar}(\boldsymbol{x})\}_{\kappa^\prime=1}^{K}\] are both weakly dependent. We state the independence here, which is needed to satisfy the conditions of Lemma~\ref{lemma:main_prod_sum_weak}. Since $\boldsymbol{z}^{\kappa^\prime\hbar-1}$ is only related to the layers before $\kappa^\prime\hbar$, while $\Delta^{\kappa^\prime\hbar}$ is only related to the layers equal to or after $\kappa^\prime\hbar$, they are independent. $\boldsymbol{z}^{\kappa^\prime\hbar-1}(\boldsymbol{x})$ and $\boldsymbol{z}^{\kappa^\prime\hbar-1}(\boldsymbol{x}^\prime)$ are also independent because the inputs are actually independent, and the same is true for $\Delta^{\kappa^\prime\hbar}(\boldsymbol{x})$ and $\Delta^{\kappa^\prime\hbar}(\boldsymbol{x}^\prime)$. Therefore, the proof can be divided into two steps as follows, followed by a conclusion.

{\bfseries Step 1. The weak dependence of $\{\boldsymbol{z}^{\kappa^\prime\hbar-1}(\boldsymbol{x})\}_{\kappa^\prime=1}^{K}.$} For the former sequence $\{\boldsymbol{z}^{\kappa^\prime\hbar-1}(\boldsymbol{x})\}_{\kappa^\prime=1}^{K}$, we prove it is a weakly dependent subsequence constructed from $\{\boldsymbol{z}^l\}_{l=0}^{L}$ by Lemma~\ref{lemma:main_weak}. According to Eq.~\eqref{eq:forward}, we have $\boldsymbol{z}^l = \phi(\hat{\mathbf{W}}^l {\boldsymbol{z}}^{l-1})$ for all $l \in [L]$. Given the well-posed $\phi$ and stable-pertinent parameter matrices, we have $\phi$ is first-order differentiable, with its derivative bounded by a certain constant $C_{\phi}$, and the inequality $ C_{\phi} \|\mathbf{W}\|_s \leq \epsilon$ holds, where $\epsilon$ is a constant that satisfies $0 < \epsilon < 1$. Thus, the conditions of Lemma~\ref{lemma:main_weak} are obviously satisfied as $\|{\partial \boldsymbol{z}^{l+\hbar}}/{\partial \boldsymbol{z}^l}\|_s \leq \epsilon^\hbar$ and ${| \boldsymbol{z}_p^{l+\hbar}\boldsymbol{z}_q^l|}/{|\boldsymbol{z}_q^l|} \leq |\boldsymbol{z}_q^l|~ \epsilon^\hbar$, where $\boldsymbol{z}_p^{l+\hbar}$ and $\boldsymbol{z}_q^l$ separately denote the $p$-th element of $\boldsymbol{z}^{l+\hbar}$ and the $q$-th element of $\boldsymbol{z}^l$. Thus, the sequence $\{\boldsymbol{z}^{\kappa^\prime\hbar-1}(\boldsymbol{x})\}_{\kappa^\prime=1}^{K}$ is weakly dependent.

{\bfseries Step 2. The weak dependence of $\{\Delta^{\kappa^\prime\hbar}(\boldsymbol{x})\}_{\kappa^\prime=1}^{K}.$} Secondly, we consider the sequence $\{\Delta^{\kappa^\prime\hbar}(\boldsymbol{x})\}_{\kappa^\prime=1}^{K}$ , where
\[
\Delta^{\kappa^\prime\hbar}(\boldsymbol{x}) = \sum_{\kappa = \kappa^\prime}^{K}\left[\prod_{i=\kappa^\prime\hbar}^{\kappa\hbar}\left(\hat{\mathbf{W}}^{i}\right)^\top\mathbf{D}^{i}(\boldsymbol{x})\right]~{\mathbf{J}^{\kappa\hbar}}^\top\quad \text{and}\quad \mathbf{D}^i(\boldsymbol{x}) = \text{diag}\{\dot{\phi}(\hat{\mathbf{W}}^i\boldsymbol{z}^{i-1}(\boldsymbol{x}))\} \ .
\]
It is observed that $\Delta^{\kappa^\prime\hbar}(\boldsymbol{x})$ is a term of the sum, normalized by the all-ones matrix ${\mathbf{J}^{\kappa\hbar}}^\top$. Although the terms in the summation are not independent, they exhibit weak dependence, which is proved later below. Thus, we conjecture that it does not contradict the conditions in Lemma~\ref{lemma:main_prod_sum_weak}. Further, in order to prove that $\{\Delta^{\kappa^\prime\hbar}(\boldsymbol{x})\}_{\kappa^\prime=1}^{K}$ is a weakly dependent sequence, it suffices to prove that 
\[
\left\{\prod_{i=\kappa^\prime\hbar}^{\kappa\hbar}\left(\hat{\mathbf{W}}^{i}\right)^\top\mathbf{D}^{i}(\boldsymbol{x}){\mathbf{J}^{\kappa\hbar}}^\top\right\}_{\kappa^\prime=1}^K
\]
is a weakly dependent sequence for any $\kappa = \kappa^\prime + 1,\kappa^\prime + 2,\cdots,K$. Also, it suffices to prove that, when $\kappa = K$, the above sequence is weakly dependent, which is the most general case. Further, the only effect of the all-ones matrix is to normalize the matrix before it. Specifically, it calculates the row sums of the matrix before it, which does not affect the weak dependence. So we only need to prove that $\{A^{\kappa^\prime\hbar}\}_{\kappa^\prime=1}^K$ is weakly dependent, where
\[A^{\kappa^\prime\hbar} = \prod_{i=\kappa^\prime\hbar}^{K\hbar}\left(\hat{\mathbf{W}}^{i}\right)^\top\mathbf{D}^{i}(\boldsymbol{x})\ .\] 
We declare that the two conditions of Lemma~\ref{lemma:main_weak} can be satisfied as 
\begin{equation}\label{eq: frist_inequality}
	\left\|{\partial A^{\kappa^\prime\hbar}}/{\partial A^{\kappa^\prime\hbar+\hbar}}\right\|_s \leq \epsilon^{\hbar}\ ,
\end{equation}
and
\begin{equation}\label{eq: second_inequality}
	\frac{|A_{pq}^{\kappa^\prime\hbar}|_{\max} ~|A_{ij}^{\kappa^\prime\hbar+\hbar}|_{\max}}{|A_{ij}^{\kappa^\prime\hbar+\hbar}|_{\max}} \leq \left({\left(R_d\right)}^{3/2} |A_{ij}^{\kappa^\prime\hbar+\hbar}|_{\max}\right) ~\epsilon^{\hbar}\ .
\end{equation}
We successively prove these two inequalities. 

{\bfseries Proof of Eq.~\eqref{eq: frist_inequality}.} The relation between $A^{\kappa^\prime\hbar+\hbar}$ and $A^{\kappa^\prime\hbar}$ can be described as
\begin{equation}\label{relation_between_A}
	A^{\kappa^\prime\hbar} = \mathbf{M} ~A^{\kappa^\prime\hbar+\hbar}
	\quad\text{with}\quad
	\mathbf{M} = \prod_{i=\kappa^\prime\hbar}^{\kappa^\prime\hbar+\hbar-1}\left(\hat{\mathbf{W}}^{i}\right)^\top\mathbf{D}^{i}(\boldsymbol{x}) \ .
\end{equation}
Firstly, by Eq.~\eqref{relation_between_A}, we have
\begin{equation}\label{last_eq}
	\quad\left\|{\partial A^{\kappa^\prime\hbar}}/{\partial A^{\kappa^\prime\hbar+\hbar}}\right\|_s = \left\|{\partial ~\mathbf{M}~A^{\kappa^\prime\hbar+\hbar}}/{\partial A^{\kappa^\prime\hbar+\hbar}}\right\|_s 
	= \left\|\mathbf{I}_{n_{K\hbar}}\otimes{\mathbf{M}}^\top \right\|_s = \left\|{\mathbf{M}}\right\|_s\ .
\end{equation}

We explain why the last step in Eq.~\eqref{last_eq} holds. Here, $\mathbf{I}_{n_{K\hbar}}$ is the identity matrix with the dimension of $n_{K\hbar}\times n_{K\hbar}$, so the term inside the norm on the left-hand side of the last equation is a block diagonal matrix, with each block being the same. Notice that the matrix norm is defined as the spectral norm here, which is equivalent to the largest singular value of the matrix. For a block diagonal matrix with all the same blocks, the spectral norm equals the largest singular value of any block. Based on this property, the last step in Eq.~\eqref{last_eq} is justified.

By the conditions in Theorem~\ref{thm:NTK_depth}, we have assumed that $\phi$ is well-posed and the weight matrices are stable-pertinent for $\phi$, so we have
\begin{equation} \label{eq: first_condition_epsilon}
	\left\|{\mathbf{M}}\right\|_s = \left\|\prod_{i=\kappa^\prime\hbar}^{\kappa^\prime\hbar+\hbar-1}\left(\hat{\mathbf{W}}^{i}\right)^\top\mathbf{D}^{i}(\boldsymbol{x})\right\|_s \leq \prod_{i=\kappa^\prime\hbar}^{\kappa^\prime\hbar+\hbar-1}\left\|\hat{\mathbf{W}}^{i}\right\|_s\left\|\mathbf{D}^{i}(\boldsymbol{x})\right\|_s \leq \prod_{i=\kappa^\prime\hbar}^{\kappa^\prime\hbar+\hbar-1}C_\phi\left\|\hat{\mathbf{W}}^{i}\right\|_s  \leq \epsilon^\hbar\ .
\end{equation}

By Eqs.~\eqref{last_eq} and~\eqref{eq: first_condition_epsilon}, Eq.~\eqref{eq: frist_inequality} is proved. 

{\bfseries Proof of Eq.~\eqref{eq: second_inequality}.} The dimensions of $A^{\kappa^\prime\hbar+\hbar}$ and $A^{\kappa^\prime\hbar}$ are finite, and we use a constant $R_d$ to denote the upper bound of the dimensions, that is, $R_d = \max\{n_{\kappa^\prime\hbar-1},n_{\kappa^\prime\hbar+\hbar-1},n_{K\hbar}\}$. According to Lemma~\ref{appendix:lemma:scale_inequalities}, we have
\begin{equation}\label{element_division}
	\frac{|A_{pq}^{\kappa^\prime\hbar}|_{\max}}{|A_{ij}^{\kappa^\prime\hbar+\hbar}|_{\max}}
	\leq \frac{\sqrt{R_d}~\lambda_{\max}(A^{\kappa^\prime\hbar})}{\frac{1}{\sqrt{R_d^2}}~{\lambda_{\max}(A^{\kappa^\prime\hbar+\hbar})}}
	\leq {\left(R_d\right)}^{3/2}~\frac{\|A^{\kappa^\prime\hbar}\|_s}{\|A^{\kappa^\prime\hbar+\hbar}\|_s}\ ,
\end{equation}
where $A_{pq}^{\kappa^\prime\hbar}$ and $A_{ij}^{\kappa^\prime\hbar+\hbar}$ separately denote any element in $A^{\kappa^\prime\hbar}$ and $A_{ij}^{\kappa^\prime\hbar+\hbar}$.

By Eq.~\eqref{relation_between_A}, we have
\[
\left\|A^{\kappa^\prime\hbar}\right\|_s = \left\|\mathbf{M}~ A^{\kappa^\prime\hbar+\hbar}\right\|_s \leq \left\|\mathbf{M}\right\|_s~\left\|A^{\kappa^\prime\hbar+\hbar}\right\|_s\ .
\]
Thus, we have
\begin{equation}\label{norm_division}
	\frac{\|A^{\kappa^\prime\hbar}\|_s}{\|A^{\kappa^\prime\hbar+\hbar}\|_s} \leq \left\|\mathbf{M}\right\|_s
	\leq \epsilon^{\hbar}\ .
\end{equation}
By substituting Eq.~\eqref{norm_division} into Eq.~\eqref{element_division}, we have
\[
\frac{|A_{pq}^{\kappa^\prime\hbar}|_{\max}}{|A_{ij}^{\kappa^\prime\hbar+\hbar}|_{\max}} \leq {\left(R_d\right)}^{3/2} ~\epsilon^{\hbar} .
\]
Moreover, we have
\begin{equation*}
	\frac{|A_{pq}^{\kappa^\prime\hbar}|_{\max}|A_{ij}^{\kappa^\prime\hbar+\hbar}|_{\max}}{|A_{ij}^{\kappa^\prime\hbar+\hbar}|_{\max}} \leq \left({\left(R_d\right)}^{3/2}~|A_{ij}^{\kappa^\prime\hbar+\hbar}|_{\max}\right) ~\epsilon^{\hbar} ,
\end{equation*}
which concludes the proof of Eq.~\eqref{eq: second_inequality}.

Since $\{|A_{ij}^{\kappa^\prime\hbar}|_{\max}\}_{\kappa^\prime=1}^K$ can be viewed as the enveloping stochastic process of $\{A_{ij}^{\kappa^\prime\hbar}\}_{\kappa^\prime=1}^K$, we here conjecture that the weak dependence between them might exhibit similarities.
Thus, by Eqs.~\eqref{eq: frist_inequality} and~\eqref{eq: second_inequality}, $\{A^{\kappa^\prime\hbar}\}_{\kappa^\prime=1}^{K}$ satisfies the conditions of Lemma~\ref{lemma:main_weak}, ensuring that it is a weakly dependent subsequence constructed from $\{A^{\kappa^\prime}\}_{\kappa^\prime=1}^{K}$. Consequently, the sequence $\{\Delta^{\kappa^\prime\hbar}(\boldsymbol{x})\}_{\kappa^\prime=1}^{K}$ is weakly dependent.  

{\bfseries Step 3. Conclusion.} To sum up the conclusions above, the sequences $\{\boldsymbol{z}^{\kappa^\prime\hbar-1}(\boldsymbol{x})\}_{\kappa^\prime=1}^{K}$ and $\{\Delta^{\kappa^\prime\hbar}(\boldsymbol{x})\}_{\kappa^\prime=1}^{K}$
are both weakly dependent, ensuring the weak dependence of the sequence $\{\text{NTK}_{(d)}^{\kappa^\prime\hbar}\}_{\kappa^\prime = 1}^{K}$. From~\citep[chap.~1.3]{doukhan2012:mixing}, it is observed that a $\beta$-mixing sequence with an exponential rate of convergence is covered by an $\alpha$-mixing sequence with a rate of $\mathcal{O}(n^{-5})$. Thus, the sequence $\{\text{NTK}_{(d)}^{\kappa^\prime\hbar}\}_{\kappa^\prime = 1}^{K}$ satisfies the conditions of Lemma~\ref{weakCLT}, that is, the Generalized Central Limit Theorem. Therefore, NTK$_{(d)}$ converges to a Gaussian distribution as $K\rightarrow +\infty$, which completes the proof of Theorem~\ref{thm:NTK_depth}. $\hfill\blacksquare$

\section{Full Proof of Theorem~\ref{thm:smallest_eigenvalue}}\label{appendix: full_proof_small} 
{\textit{Proof.}} For simplicity, we force $n = n_o =  n_\hbar = n_{2\hbar}  = \cdots = n_{K\hbar}$, while keeping $d = n_0$ independent from $n$. Thus, we can omit the all-ones matrices $\mathbf{J}^{\kappa\hbar}$ in Eq.~\eqref{eq:shortcut}. Also, we neglect the scaling constant $1/\sqrt{M_{\boldsymbol{z}}}$, which is barely related to the input dimension $d$.
	
Firstly, we recall two inequalities about the smallest eigenvalues of matrices as follows
\begin{equation}\label{main_lambda_min}
	\lambda_{\min}(A+B) \geq \lambda_{\min}(A) + \lambda_{\min}(B)\quad\text{and}\quad
	\lambda_{\min}(A\circ B) \geq \lambda_{\min}(A){\min}_{i\in[n]}B_{ii}\ ,
\end{equation}
where the first inequality holds for Hermitian matrices~\citep{weyl1949inequalities} and the second one holds for p.s.d. matrices~\citep{Schur1911}.
	
We care about the lower bound on the smallest eigenvalue of NTK$_{(d)}$, which is composed of $N\times N$ block matrices, where the $(i,j)$-th block is NTK$_{(d)}(i,j)$ defined by Eq.~\eqref{main_NTK_def}. Simply from the definition, it follows that all of these matrices are real symmetric, and hence a special case of Hermitian matrices. Thus, by Eqs.~\eqref{main_NTK_sum} and~\eqref{main_lambda_min}, it holds that
\begin{equation}\label{main_eq:sum_large}
	\lambda_{\min}(\text{NTK}_{(d)}) = \lambda_{\min}\left(\sum_{\kappa^\prime = 1}^{K}\text{NTK}_{(d)}^{\kappa^\prime\hbar}\right) \geq \sum_{\kappa^\prime = 1}^{K}\lambda_{\min}\left(\text{NTK}_{(d)}^{\kappa^\prime\hbar}\right) \geq \lambda_{\min}\left(\text{NTK}_{(d)}^{K\hbar}\right)\ ,
\end{equation}
where the last inequality follows from the fact that all the matrices in Eq.~\eqref{main_eq:sum_large} are p.s.d. matrices. Now it suffices to lower bound $\lambda_{\min}(\text{NTK}_{(d)}^{K\hbar})$.
	
Let ${\mathbf{Z}^{K\hbar}}^\top$ denote the column vector vec$(\boldsymbol{z}^{K\hbar}(\boldsymbol{x}_i)^\top)_{i=1}^N$. Let ${\boldsymbol{\Delta}^{K\hbar}}^\top$ denote the column vector vec$(\Delta^{K\hbar}(\boldsymbol{x}_i)^\top)_{i=1}^N$. Then, by Eq.~\eqref{main_NTK_exp}, NTK$_{(d)}^{K\hbar}$ can be written as the form of a Hadamard product by
\begin{equation}\label{NTK_element}
	\text{NTK}_{(d)}^{K\hbar} = \left({\mathbf{Z}^{K\hbar}}^\top {\mathbf{Z}^{K\hbar}}\right) \circ \left({\boldsymbol\Delta^{K\hbar}}^\top{\boldsymbol\Delta^{K\hbar}}\right)\ .
\end{equation}
We offer additional clarifications on the Hadamard product in Eq.~\eqref{NTK_element}. Notice that ${\mathbf{Z}^{K\hbar}}^\top {\mathbf{Z}^{K\hbar}}$ is a $N\times N$ matrix whose entries are scalars, while ${\boldsymbol\Delta^{K\hbar}}^\top{\boldsymbol\Delta^{K\hbar}}$ is a $N\times N$ matrix whose entries are block matrices. The original computing result of NTK$_{(d)}^{K\hbar}$ is a $N\times N$ matrix whose entries are $n\times n$ matrices given by
\[\text{NTK}_{(d)}^{K\hbar}(i,j) = \left({\mathbf{Z}^{K\hbar}}^\top {\mathbf{Z}^{K\hbar}}\right)_{ij}\left({\boldsymbol\Delta^{K\hbar}}^\top{\boldsymbol\Delta^{K\hbar}}\right)_{ij}\ ,
\]
which corresponds to the multiplication of a scalar and a matrix. Moreover, multiplying a scalar by a matrix is equivalent to: (1) firstly replacing the scalar with an $n\times n$ matrix where every entry equals the very scalar; and (2) applying the Hadamard product between the two matrices. Thus, NTK$_{(d)}^{K\hbar}$ can be written by a Hadamard product in Eq.~\eqref{NTK_element}.
	
Then, by invoking Eq.~\eqref{main_lambda_min} into Eq.~\eqref{NTK_element}, we have
\begin{equation}\label{main_apart_z_delta}
	\lambda_{\min}\left(\text{NTK}_{(d)}^{K\hbar}\right) \geq \lambda_{\min}\left({\mathbf{Z}^{K\hbar}}^\top {\mathbf{Z}^{K\hbar}}\right)\min_{i\in[n\times N]}\left({\boldsymbol\Delta^{K\hbar}}^\top{\boldsymbol\Delta^{K\hbar}}\right)_{ii}\ .
\end{equation}
Notice that Eq.~\eqref{main_apart_z_delta} consists of the product of two components, the term of $\mathbf{Z}$ and the term of $\boldsymbol\Delta$. For the term of $\boldsymbol\Delta$, we have two observations that make it unrelated to the lower bound of $\lambda_{\min}(\text{NTK}_{(d)}^{K\hbar})$. (1) It is unrelated to the input dimension $d$; and (2) it is a rare event that the term of $\boldsymbol\Delta$ reduces to zero. To clarify the second observation, we consider the event when the term of $\boldsymbol\Delta$ reduces to zero, taking the first $n\times n$ block matrix of ${\boldsymbol\Delta^{K\hbar}}^\top{\boldsymbol\Delta^{K\hbar}}$ for example. The matrix has $n$ diagonal elements, whose $i$-th diagonal element is given by
\[\sum_{j=1}^{n} \left({\boldsymbol\Delta}^{K\hbar}(\boldsymbol{x}_1)\right)_{ji}^2 =  \sum_{j=1}^{n} {\left({\left(\hat{\mathbf{W}}^{K\hbar}\right)}^\top~\mathbf{D}^{K\hbar}(\boldsymbol{x}_1)\right)}_{ji}^2 \quad \text{for}\quad i\in[n]\ .
\]
If one of the diagonal elements, say the $1$-th diagonal element, becomes zero, it implies that the entire first column of ${(\hat{\mathbf{W}}^{K\hbar})}^\top\mathbf{D}^{K\hbar}(\boldsymbol{x}_1)$ must be zero. Furthermore, because the activation here is Leaky ReLU, $\mathbf{D}^{K\hbar}(\boldsymbol{x}_1)$ is a diagonal matrix whose entries are larger than zero. Thus, the entire first row of $\hat{\mathbf{W}}^{K\hbar}$ must be zero, which is a rare event, and can be easily avoided by suitable initialization.
	
Now the problem lies in giving the lower bound of $\lambda_{\min}({\mathbf{Z}^{K\hbar}}^\top {\mathbf{Z}^{K\hbar}})$. We firstly introduce two lemmas about the Hermite expansion of the activation and the data distribution.
	
\begin{lemma}\label{lemma: Hermite_expand}For any even integer $r \geq 2$, the Leaky ReLU $\phi(x) = max(x,\alpha x)$ has the Hermite expansion with the following form,
	\begin{equation}\label{eq: hermite}
		\mu_r(\phi) = \frac{1-\alpha}{\sqrt{2\pi}}{(-1)}^{\frac{r-2}{2}}\frac{(r-3)!!}{\sqrt{r!}}\ .
	\end{equation}
\end{lemma}
Lemma~\ref{lemma: Hermite_expand} can be simply proved by standard calculations. For a special case, if we set $\alpha = 0$, then Eq.~\eqref{eq: hermite} turns into the Hermite expansion of ReLU, which has been exploited in previous research~\citep{pmlr-v139-nguyen21g}.
	
\begin{lemma}\label{eq:data_distribution_deduce}~\citep[Lemma 8]{Zhang2024:deep}
	Let $\left\{\boldsymbol{x}_i\right\}_{i=1}^{N}$ be a set of i.i.d. data points from $P_X$. If $P_X$ is a centered Gaussian distribution, then, for $i \neq j$, it holds with probability of at least \(1-N\exp({-\Omega(d)})-N^2\exp({-\Omega(dN^{-2/(r-0.5)}))}\) that
		\[\|\boldsymbol{x}_i\|_2 = \Theta(\sqrt{d}) \quad\text{and}\quad |\langle\boldsymbol{x}_i,\boldsymbol{x}_j\rangle|^r \leq dN^{-1/r-0.5}\ .\]
\end{lemma}
Lemma~\ref{eq:data_distribution_deduce} can be regarded as a corollary applicable to our well-scaled distribution $P_X$, as the latter can be a centered Gaussian distribution in our setting.
	
Now, we formally give the lower bound of $\lambda_{\min}({\mathbf{Z}^{K\hbar}}^\top {\mathbf{Z}^{K\hbar}})$. First, by Eq.~\eqref{eq:forward}, we have ${\mathbf{Z}^{K\hbar}}^T{\mathbf{Z}^{K\hbar}} = \phi(\mathbf{W}^{K\hbar}\mathbf{Z}^{K\hbar-1})^\top\phi(\mathbf{W}^{K\hbar}\mathbf{Z}^{K\hbar-1})$. Then, by using Eq.~\eqref{eq: hermite}, we can bound $\lambda_{\min}({\mathbf{Z}^{K\hbar}}^T{\mathbf{Z}^{K\hbar}})$ in terms of $\lambda_{\min}(({\mathbf{Z}^{K\hbar-1}}^{(r)})^\top({\mathbf{Z}^{K\hbar-1}}^{(r)}))$ for any even integer $r \geq 2$, where $(\cdot)^{(r)}$ denotes the $r$-th Khatri Rao power. By recursively applying this argument, it suffices to bound $\lambda_{\min}(({\mathbf{X}}^{(r)})^\top({\mathbf{X}}^{(r)}))$, where ${\mathbf{X}}^\top$ denotes the column vector vec$(\boldsymbol{x}_i^\top)_{i=1}^N$. In what follows, we focus on the final step, namely the transition from $\mathbf{Z}^1$ to $\mathbf{X}$.
	
\begin{equation}\label{lower_last_step}
	\begin{aligned}
		\lambda_{\min}\left({\mathbf{Z}^1}^\top{\mathbf{Z}^1}\right) &= \lambda_{\min}\left(\phi\left(\mathbf{W}^{1}\mathbf{X}\right)^\top\phi\left(\mathbf{W}^{1}\mathbf{X}\right)\right)\\
		& \geq \mu_r(\phi)^2\lambda_{\min}\left(\left({\mathbf{X}}^{(r)}\right)^\top\left({\mathbf{X}}^{(r)}\right)\right)\\
		& \geq \mu_r(\phi)^2\left(\min_{i\in[N]}\left\|\boldsymbol{x}_i\right\|_2^{2r} - (N-1)\max_{j \neq i}|\langle\boldsymbol{x}_i,\boldsymbol{x}_j\rangle|^r\right)\\
		& \geq \mu_r(\phi)^2\Omega(d)\ .
	\end{aligned}
\end{equation}
	
In Eq.~\eqref{lower_last_step}, the second inequality holds by Gershgorin Circle Theorem~\citep{SALAS199915}, and the third inequality holds by Lemma~\ref{eq:data_distribution_deduce}. Thus, we obtain the lower bound. $\hfill\blacksquare$

\section{Proof of Theorem~\ref{thm:constant}}\label{proof_constant}
{\bf Statement of Theorem~\ref{thm:constant}}~ Provided the shortcut-related network defined by Eqs.~\eqref{eq:forward} and~\eqref{eq:shortcut}, if the following conditions hold
\begin{itemize}
	\item[(1)] $\phi$ is well-posed and satisfies the Lipschitz condition with $\|\phi(\boldsymbol{x}) - \phi(\boldsymbol{y})\|_s \leq L\|\boldsymbol{x}-\boldsymbol{y}\|_s$,
	\item[(2)] the weight matrices are stable-pertinent for $\phi$, that is, $\mathbf{W}^l \in SP(\phi)$ for $\forall~ l\in [L]$,
	\item[(3)] we consider any T such that the integral $\int_0^T\|d_t\|_{\mathcal{D}}\, dt$ remains stochastically bounded,
\end{itemize}
given a constant $\lambda$ satisfying $0 < \lambda<1$, as $K\rightarrow+\infty$ and $n\rightarrow+\infty$ with the ratio of $n=\Omega(K^{2/\lambda})$, it holds uniformly that $\text{NTK}_{(d)}(t) \rightarrow \text{NTK}_{(d)}(0)$ for $t\in[0,T]$.

{\textit{Proof.}} For simplicity, we force $n = n_o = n_0 = n_\hbar = \cdots = n_{K\hbar}$. Thus, we can omit the all-ones matrices $\mathbf{J}^{\kappa\hbar}$ in Eq.~\eqref{eq:shortcut}, which are previously employed to normalize the dimensions. For simplicity, we denote the limit in Theorem~\ref{thm:constant} by $(n,K)\rightarrow(+\infty,+\infty)_\lambda$.

In order to prove that NTK$_{(d)}(t)\rightarrow$ NTK$_{(d)}(0)$, by Eqs.~\eqref{sum_of_NTK_kappa_hbar} and~\eqref{expanded_form}, it suffices to prove that for any $\kappa^\prime\in [K]$, it holds $\boldsymbol{z}^{\kappa^\prime\hbar}(t) \rightarrow \boldsymbol{z}^{\kappa^\prime\hbar}(0)$ and $\Delta^{\kappa^\prime\hbar}(t) \rightarrow \Delta^{\kappa^\prime\hbar}(0)$ as $(n,K)\rightarrow(+\infty,+\infty)_\lambda$.

We start the proof with a lemma as follows.

\begin{lemma}\label{lemma: W_minus_bound}
	With the settings of Theorem~\ref{thm:constant}, for any $\kappa^\prime\in[K]$, we have
	\[
	\underset{(n,K)\rightarrow(+\infty,+\infty)_\lambda}{\text{lim}}\frac{
		\partial \left\|{\hat{\mathbf{W}}^{\kappa^\prime\hbar}(t) - \hat{\mathbf{W}}^{\kappa^\prime\hbar}(0)} \right\|_{s}}{\partial t} = 0 ,
	\]
	which means that $\hat{\mathbf{W}}^{\kappa^\prime\hbar}(t) \rightarrow \hat{\mathbf{W}}^{\kappa^\prime\hbar}(0)$ in the limit of $(n,K)\rightarrow(+\infty,+\infty)_\lambda$.
\end{lemma}
The key idea of proving Lemma~\ref{lemma: W_minus_bound} refers to that of Jacot et al.~\citep[Theorem 2]{WideNTK2021}. By Eq.~\eqref{eq: dt}, and notice that $\boldsymbol{z}^{\kappa^\prime\hbar}$ influences $\hat{\mathbf{W}}^{\kappa^\prime\hbar}$ the most, we have
\begin{equation}\label{eq: w_exponential}
	\frac{
		\partial \left\|{\hat{\mathbf{W}}^{\kappa^\prime\hbar}(t) - \hat{\mathbf{W}}^{\kappa^\prime\hbar}(0)} \right\|_{s}}{\partial t} \leq  \left\|\frac{
		{\partial\hat{\mathbf{W}}^{\kappa^\prime\hbar}(t)} }{\partial t}\right\|_{s} \leq \frac{1}{\sqrt{n~M_{\boldsymbol{z}}}}\ \left\|\boldsymbol{z}^{\kappa^\prime\hbar}(t)\right\|_{\mathcal{D}}\left\|d_t\right\|_{\mathcal{D}}\ .
\end{equation}
By the conditions that $\int_0^T\|d_t\|_{\mathcal{D}}\, dt$ stays bounded, and the output of the $l_{\kappa^\prime\hbar}$-th layer stays bounded during training, Lemma~\ref{lemma: W_minus_bound} holds obviously.

We then prove step by step that in the limit of $(n, K)\rightarrow(+\infty,+\infty)_\lambda$, the change of $\boldsymbol{z}^{\kappa\hbar}$ and that of $\Delta^{\kappa\hbar}$ are little related to the NTK$_{(d)}^{\kappa\hbar}$ kernel.

{\bf Step 1. Proving that $\boldsymbol{z}^{\kappa^\prime\hbar}$ changes little}

We first assume that only $\boldsymbol{z}^{\kappa^\prime\hbar}$ changes, while the other $\boldsymbol{z}^{i\hbar}~(i \neq \kappa^\prime)$ remains invariant, that is, $\boldsymbol{z}^{i\hbar}(t) \rightarrow \boldsymbol{z}^{i\hbar}(0)~ (i \neq \kappa^\prime)$ as $(n,K)\rightarrow(+\infty,+\infty)_\lambda$. In this context, as $(n,K)\rightarrow(+\infty,+\infty)_\lambda$, by the Lipschitz condition of the activation $\phi$, we have
\begin{align}\label{z_change_constant}
	\left\|{\boldsymbol{z}^{\kappa^\prime\hbar}(t) - \boldsymbol{z}^{\kappa^\prime\hbar}(0)} \right\|_{s} &= \left\| \phi\left(\hat{\mathbf{W}}^{\kappa^\prime\hbar}(t)\boldsymbol{z}^{\kappa^\prime\hbar-1}(t)\right)  - \phi\left(\hat{\mathbf{W}}^{\kappa^\prime\hbar}(0)\boldsymbol{z}^{\kappa^\prime\hbar-1}(0)\right)\right\|_{s}\nonumber \\
	&\leq L\left\|\hat{\mathbf{W}}^{\kappa^\prime\hbar}(t)\boldsymbol{z}^{\kappa^\prime\hbar-1}(t) - \hat{\mathbf{W}}^{\kappa^\prime\hbar}(0)\boldsymbol{z}^{\kappa^\prime\hbar-1}(0)  \right\|_s \nonumber \\
	&\leq L~\left\|\boldsymbol{z}^{\kappa^\prime\hbar}(0)\right\|_s \left\|{\hat{\mathbf{W}}^{\kappa^\prime\hbar}(t) - \hat{\mathbf{W}}^{\kappa^\prime\hbar}(0)} \right\|_{s}\ .
\end{align}
Thus, by Lemma~\ref{lemma: W_minus_bound}, we have that $\boldsymbol{z}^{\kappa^\prime\hbar}(t) \rightarrow \boldsymbol{z}^{\kappa^\prime\hbar}(0)$ in the limit of $(n,K)\rightarrow(+\infty,+\infty)_\lambda$.

Then we consider that $\boldsymbol{z}^{\kappa^\prime\hbar}$ changes for any $\kappa^\prime \in [K]$. Notice that our focus is indeed on the change of NTK$_{(d)}$, and studying the variation of $\boldsymbol{z}^{\kappa^\prime\hbar}$ serves as an intermediate step to analyze the corresponding change in NTK$_{(d)}$. By Eqs.~\eqref{eq: w_exponential} and~\eqref{z_change_constant}, the change of each $\boldsymbol{z}^{\kappa^\prime\hbar}$ is small, with an exponential speed of $\mathcal{O}(1/\sqrt{n})$. However, the contributions of $\boldsymbol{z}$ to NTK$_{(d)}$ appear in a summation form across all layers with shortcuts, which is polynomial. Thus, in total, the influence of the change of $\boldsymbol{z}^{\kappa^\prime\hbar}$ on NTK$_{(d)}$ reduces to zero as $(n,K)\rightarrow(+\infty,+\infty)_\lambda$.  

{\bf Step 2. Proving that ${\Delta}^{\kappa^\prime\hbar}$ changes little}

We first provide a lemma here, which is proved in Appendix~\ref{proof_lemma_wd} below.
\begin{lemma}\label{lemma: delta_minus_bound}
	With the settings of Theorem~\ref{thm:constant}, it holds that
	\[
	\resizebox{0.99\textwidth}{!}{$
		\left\|{\Delta}^{\kappa^\prime\hbar}(t) - {\Delta}^{\kappa^\prime\hbar}(0)\right\|_{s} \leq \frac{2K+4}{\sqrt{n}} + C_{\hat{\mathbf{W}},{\mathbf{D}}} ~\left\{ \left\| {\mathbf{D}}^{K\hbar}(t) - {\mathbf{D}}^{K\hbar}(0) \right\|_s +   \left\|  \hat{\mathbf{W}}^{K\hbar} (t) -  \hat{\mathbf{W}}^{K\hbar} (0) \right\|_s\right\} \ ,
		$}
	\]
	where $C_{\hat{\mathbf{W}},{\mathbf{D}}}$ is a constant related to $\hat{\mathbf{W}}$ and ${\mathbf{D}}$.
\end{lemma}
According to Lemma~\ref{lemma: W_minus_bound} and Lemma~\ref{lemma: delta_minus_bound}, in order to prove that ${\Delta}^{\kappa^\prime\hbar}(t) \rightarrow {\Delta}^{\kappa^\prime\hbar}(0)$ as $(n,K)\rightarrow(+\infty,+\infty)_\lambda$, it suffices to prove that $ {\mathbf{D}}^{K\hbar}(t)\rightarrow{\mathbf{D}}^{K\hbar}(0)$ as $(n,K)\rightarrow(+\infty,+\infty)_\lambda$. Since it is proved that $\boldsymbol{z}^{\kappa^\prime\hbar}(t) \rightarrow \boldsymbol{z}^{\kappa^\prime\hbar}(0)$ and $\hat{\mathbf{W}}^{\kappa^\prime\hbar}(t)\rightarrow\hat{\mathbf{W}}^{\kappa^\prime\hbar}(0)$ as $(n,K)\rightarrow(+\infty,+\infty)_\lambda$, we have $ {\mathbf{D}}^{K\hbar}(t)\rightarrow{\mathbf{D}}^{K\hbar}(0)$ in the same limit. Furthermore, we have $(2K+4)/\sqrt{n}\rightarrow 0$, which completes the proof of Theorem~\ref{thm:constant}. \hfill$\blacksquare$

\section{Proof of Lemma~\ref{lemma: delta_minus_bound}}\label{proof_lemma_wd}

{\textit{Proof.}}  By the conditions that $\phi$ is well-posed and the weight matrices are stable-pertinent for $\phi$, it holds that
\begin{align}\label{eq:delta_minus_s_norm}
	\left\| {\Delta}^{\kappa^\prime\hbar} - {\Delta}^{K\hbar} \right\|_s &\leq \left\|{\Delta}^{\kappa^\prime\hbar}\right\|_s + \left\|{\Delta}^{K\hbar}\right\|_s \nonumber \\
	&=  \left\|\sum_{\kappa = \kappa^\prime}^{K}\frac{1}{n^{(\kappa-\kappa^\prime+1)/2}} \left[\prod_{i=\kappa^\prime\hbar}^{\kappa\hbar}{\mathbf{W}^{i}}^\top{\mathbf{D}}^{i}\right]\right\|_s + \frac{1}{n^{1/2}}\left\|{{\mathbf{W}}^{K\hbar}} ^\top{\mathbf{D}}^{K\hbar}\right\|_s\nonumber  \\
	&\leq \sum_{\kappa = \kappa^\prime}^{K}\frac{\epsilon^{(\kappa-\kappa^\prime)\hbar +1}}{n^{(\kappa-\kappa^\prime+1)/2}} + \frac{1}{n^{1/2}}
	\leq \frac{K-\kappa^\prime+1}{n^{1/2}} + \frac{1}{n^{1/2}} \leq \frac{K+2}{\sqrt{n}}\ .
\end{align}
By the triangle inequality of norms and by Eq.~\eqref{eq:delta_minus_s_norm}, we have

\begin{align}
	\scalebox{0.98}{$
		\begin{aligned}\label{eq: delta_t_minus_0}
			\left\|{\Delta}^{\kappa^\prime\hbar}(t) - {\Delta}^{\kappa^\prime\hbar}(0)\right\|_{s} 
			&\leq \left\|{\Delta}^{\kappa^\prime\hbar}(t) - {\Delta}^{K\hbar}(t)\right\|_{s} + \left\|{\Delta}^{K\hbar}(t) - {\Delta}^{K\hbar}(0)\right\|_{s} + \left\|{\Delta}^{K\hbar}(0) - {\Delta}^{\kappa^\prime\hbar}(0)\right\|_{s} \\
			&\leq \frac{2K+4}{\sqrt{n}} + \left\|{\Delta}^{K\hbar}(t) - {\Delta}^{K\hbar}(0)\right\|_{s} \\
			&= \frac{2K+4}{\sqrt{n}} + \left\| \left( \hat{\mathbf{W}}^{K\hbar} \right)^\top(t)~{\mathbf{D}}^{K\hbar}(t) - \left( \hat{\mathbf{W}}^{K\hbar} \right)^\top(0)~{\mathbf{D}}^{K\hbar}(0)  \right\|_s\ .
		\end{aligned}
		$}
\end{align}
For the second term in the last line, it holds
\begin{align}
	\scalebox{0.88}{$
		\begin{aligned}\label{eq: delta_t_minus_0_scale}
			&\left\| \left( \hat{\mathbf{W}}^{K\hbar} \right)^\top(t)~{\mathbf{D}}^{K\hbar}(t) - \left( \hat{\mathbf{W}}^{K\hbar} \right)^\top(0)~{\mathbf{D}}^{K\hbar}(0)  \right\|_s \\
			&\leq \left\|\left( \hat{\mathbf{W}}^{K\hbar} \right)^\top(t)~{\mathbf{D}}^{K\hbar}(t) - \left( \hat{\mathbf{W}}^{K\hbar} \right)^\top(t)~{\mathbf{D}}^{K\hbar}(0) \right\|_s + \left\|\left( \hat{\mathbf{W}}^{K\hbar} \right)^\top(t)~{\mathbf{D}}^{K\hbar}(0) - \left( \hat{\mathbf{W}}^{K\hbar} \right)^\top(0)~{\mathbf{D}}^{K\hbar}(0) \right\|_s \\
			&\leq \left\|  \hat{\mathbf{W}}^{K\hbar} (t) \right\|_s \left\| {\mathbf{D}}^{K\hbar}(t) - {\mathbf{D}}^{K\hbar}(0) \right\|_s +   \left\|  \hat{\mathbf{W}}^{K\hbar} (t) -  \hat{\mathbf{W}}^{K\hbar} (0) \right\|_s  \left\| {\mathbf{D}}^{K\hbar}(0) \right\|_s \\
			&\leq C_{\hat{\mathbf{W}},{\mathbf{D}}}  \left\{ \left\| {\mathbf{D}}^{K\hbar}(t) - {\mathbf{D}}^{K\hbar}(0) \right\|_s +   \left\|  \hat{\mathbf{W}}^{K\hbar} (t) -  \hat{\mathbf{W}}^{K\hbar} (0) \right\|_s\right\}\ ,
		\end{aligned}
		$}
\end{align}
where $C_{\hat{\mathbf{W}},{\mathbf{D}}}$ is a constant satisfying 
\[
\max\left\{ \left\|  \hat{\mathbf{W}}^{K\hbar} (t) \right\|_s, ~\left\| {\mathbf{D}}^{K\hbar}(0) \right\|_s\right\} \leq C_{\hat{\mathbf{W}},\mathbf{D}}\ .
\]
Notice that this bound holds simply under the assumption that the norms of weights and gradients remain bounded throughout training. By combining Eqs.~\eqref{eq: delta_t_minus_0} and ~\eqref{eq: delta_t_minus_0_scale}, we conclude the proof of Lemma~\ref{lemma: delta_minus_bound}. $\hfill\blacksquare$

\end{document}